\documentclass[10pt,twocolumn,letterpaper]{article}

\usepackage{cvpr}
\usepackage{times}
\usepackage{epsfig}
\usepackage{epstopdf}
\usepackage{graphicx}
\usepackage{amsmath}
\usepackage{amssymb}
\usepackage{url}
\usepackage{booktabs}
\usepackage{amsfonts}
\usepackage{nicefrac}
\usepackage{microtype}
\usepackage{algorithm}
\usepackage{algorithmic}
\usepackage[tight,footnotesize]{subfigure}

\usepackage[pagebackref=true,breaklinks=true,letterpaper=true,colorlinks,bookmarks=false]{hyperref}

 \cvprfinalcopy

\newtheorem{defn}{Definition}[section]
\begin{document}

\title{Decorrelated Batch Normalization}

\author{Lei Huang$^{\dag\ddag}$\thanks{This work was mainly done while Lei Huang was a visiting student at the
University of Michigan.} \quad Dawei Yang$^{\ddag}$ \quad  Bo Lang$^{\dag}$ \quad  Jia Deng $^{\ddag}$\\
$^{\dag}$State Key Laboratory of Software Development Environment, Beihang University, P.R.China\\
$^{\ddag}$University of Michigan, Ann Arbor\\
}

\maketitle
\newcommand{\TODO}[1]{\textcolor{red}{TODO: }\textcolor{red}{\emph{#1}}}

\begin{abstract}
Batch Normalization (BN) is capable of accelerating the training of
deep models by centering and scaling activations within mini-batches.
In this work, we propose
Decorrelated Batch Normalization (DBN), which not just centers and scales activations but
whitens them. We explore multiple whitening
techniques, and find that PCA whitening causes a problem we call
\emph{stochastic axis swapping}, which is detrimental to learning. We
show that ZCA whitening does not suffer from this problem, permitting successful
learning. DBN retains the desirable
qualities of BN and further improves BN's optimization efficiency and
generalization ability.  We design comprehensive experiments to show
that DBN can improve the performance of BN on multilayer perceptrons
and convolutional neural networks. Furthermore, we consistently
improve the accuracy of residual networks on CIFAR-10, CIFAR-100, and
ImageNet.
\end{abstract}

\section{Introduction}
\label{sec_intro}

Batch Normalization \cite{2015_ICML_Ioffe} is a technique for accelerating deep network
training. Introduced by Ioffe and Szegedy, it has been widely used in a variety of state-of-the-art
systems \cite{2015_CVPR_He,2015_CoRR_Szegedy,
2016_CoRR_He,2016_CoRR_Zagoruyko,2016_CoRR_Szegedy,2016_CoRR_Huang_a}.  Batch Normalization works by
standardizing the activations of a deep network within a mini-batch---transforming the
output of a layer, or equivalently the input to the next layer, to have a zero mean and unit
variance. Specifically, let $\{x_{i} \in \mathbb{R}, i=1,2,\ldots,m \}$ be the original outputs of
a single neuron on $m$ examples in a mini-batch.  Batch Normalization produces the
transformed outputs
{\setlength\abovedisplayskip{6pt}
\setlength\belowdisplayskip{6pt}
\begin{equation}
\label{bn_forward}
\hat{x}_i =  \gamma \frac{x_i - \mu}{\sqrt{\sigma^2 + \epsilon}}  + \beta,
\end{equation}
}
\hspace{-0.05in}where $\mu = \frac{1}{m}\sum_{j=1}^m x_j$ and $\sigma^2 = \frac{1}{m} \sum_{j=1}^m (x_j
-\mu)^2$ are the mean and variance of the mini-batch, $\epsilon>0$ is a small number to
prevent numerical instability, and $\gamma, \beta$ are extra learnable parameters.
Crucially, during training, Batch Normalization is part of both the inference
computation (forward pass) as well as the gradient computation (backward
pass).  Batch Normalization can be inserted extensively
into a network, typically between a linear mapping and a nonlinearity.

Batch Normalization was motivated by the well-known fact that
whitening inputs (\ie centering, decorrelating, and scaling) speeds up
training \cite{1998_NN_Yann}.  It has been shown that better
conditioning of the covariance matrix of the input leads to better
conditioning of the Hessian in updating the weights, making the
gradient descent updates closer to Newton updates
\cite{1998_NN_Yann,2011_NIPS_Wiesler}.  Batch Normalization exploits
this fact further by seeking to whiten not only the input to the first
layer of the network, but also the inputs to each internal layer in the
network.  But instead of whitening, Batch Normalization only performs
standardization. That is, the activations are centered and scaled, but
not decorrelated.  Such a choice was justified in
\cite{2015_ICML_Ioffe} by citing the cost and differentiability of
whitening, but no actual attempts were made to derive or experiment
with a whitening operation.

While standardization has proven effective for Batch Normalization, it remains an
interesting question whether full whitening---adding decorrelation to Batch
Normalization---can help further. Conceptually, there are clear cases where whitening
is beneficial. For example, when the activations are close to being perfectly\footnote{For example, in 2D,
this means all points lie close to the line $y = x$ and Batch Normalization does not
change the shape of the distribution.} correlated,
standardization barely improves the conditioning of the covariance matrix, whereas
whitening remains effective. In addition, prior work has shown that decorrelated
activations result in better
features~\cite{1961_Barlow_possible,1992_NC_Schmidhuber,2009_NIPS_Bengio} and better
generalization~\cite{2016_ICLR_Cogswell,2016_ICDM_Xiong}, suggesting room for further
improving Batch Normalization.

In this paper, we propose Decorrelated Batch Normalization, in which we whiten the activations of each layer
within a mini-batch. Let $\mathbf{x}_i \in \mathbb{R}^d$ be the input to a layer for the $i$-th example in a
mini-batch of size $m$.  The whitened input is given by
{\setlength\abovedisplayskip{6pt}
\setlength\belowdisplayskip{6pt}
\begin{equation}
\hat{\mathbf{x}}_i = \Sigma^{-\frac{1}{2}}(\mathbf{x}_i - \mathbf{\mu}),
\end{equation}
}
\hspace{-0.05in}where $\mathbf{\mu}=\frac{1}{m} \sum_{j=1}^{m} \mathbf{x}_j$ is the mini-batch mean and
$\Sigma =\frac{1}{m} \sum_{j=1}^{m}
(\mathbf{x}_j-\mathbf{\mu})(\mathbf{x}_j-\mathbf{\mu})^T$ is the mini-batch covariance matrix.

Several questions arise for implementing Decorrelated Batch
Normalization.  One is how to perform back-propagation, in particular,
how to back-propagate through the inverse square root of a matrix
(\ie $\partial\Sigma^{-\frac{1}{2}}/\partial \Sigma$), whose key step
is an eigen decomposition. The differentiability of this matrix
transform was one of the reasons that whitening was not pursued in the
Batch Normalization paper \cite{2015_ICML_Ioffe}. Desjardins \etal
\cite{2015_NIPS_Desjardins} whiten the activations but avoid
back-propagation through it by treating the mean $\mathbf{\mu}$ and
the whitening matrix $\Sigma^{-\frac{1}{2}}$ as model parameters,
rather than as functions of the input examples. However, as has been
pointed out~\cite{2015_ICML_Ioffe,2017_NIPS_Ioffe}, doing so may lead
to instability in training.

In this work, we decorrelate the activations and perform proper back-propagation
during training. We achieve this by using the fact that eigen-decomposition is
differentiable, and its derivatives can be obtained using matrix differential calculus, as
shown by prior work~\cite{2008_AD_Giles, 2015_ICCV_Ionescu}. We build upon these existing
results and derive the back-propagation updates for Decorrelated Batch Normalization.

\label{para_1_6}
Another question is, perhaps surprisingly, the choice of how to
compute the whitening matrix $\Sigma^{-\frac{1}{2}}$.  The whitening
matrix is not unique because a whitened input stays whitened after an
arbitrary rotation~\cite{2017_AS_Kessy}.  It turns out that PCA
whitening, a standard choice~\cite{2015_NIPS_Guillaume}, does not
speed up training at all and in fact inflicts significant harm. The
reason is that PCA whitening works by performing rotation followed by
scaling, but the rotation can cause a problem we call \emph{stochastic
  axis swapping}, which, as will be discussed in Section
\ref{para_3_1}, in effect randomly permutes the neurons of a layer
for each batch.
Such permutation can drastically change the data representation from
one batch to another to the extent that training never converges.

To address this stochastic axis swapping issue, we discover that it is critical to use ZCA
whitening~\cite{1997_Vr_Bell,2017_AS_Kessy}, which rotates the PCA-whitened activations back
such that the distortion of the original activations is minimal. We show through
experiments that the benefits of decorrelation are only observed with the additional
rotation of ZCA whitening.

A third question is the amount of whitening to perform. Given a
particular batch size, DBN may not have enough samples to obtain a
suitable estimate for the full covariance matrix.  We thus control the
extent of whitening by decorrelating smaller groups of activations
instead of all activations together. That is, for an output of
dimension $d$, we divide it into groups of size $k_G < d$ and apply
whitening within each group. This strategy has the added benefit of
reducing the computational cost of whitening from $O(d^2 \max(m,d))$
to $O(mdk_G)$, where $m$ is the mini-batch size.

We conduct extensive experiments on multilayer perceptrons and convolutional neural
networks, and show that Decorrelated Batch Normalization (DBN) improves upon the
original Batch Normalization (BN) in terms of training speed and generalization performance.
In particular, experiments demonstrate that using DBN can consistently improve the performance of residual networks
\cite{2015_CVPR_He,2016_CoRR_He,2016_CoRR_Zagoruyko} on CIFAR-10, CIFAR-100
~\cite{2009_TR_Alex} and ILSVRC-2012~\cite{2009_ImageNet}.

\section{Related Work}
\label{sec_relatedWork}

Normalized activations \cite{2012_NN_Gregoire, 2014_ICASSP_Wiesler}
and gradients \cite{1998_Schraudolph, 2012_AISTATS_Raiko} have long
been known to be beneficial for training neural networks.  Batch
Normalization~\cite{2015_ICML_Ioffe} was the first to perform
normalization per mini-batch in a way that supports
back-propagation. One drawback of Batch Normalization, however, is
that it requires a reasonable batch size to estimate the mean and
variance, and is not applicable when the batch size is very small. To
address this issue, Ba \etal \cite{2016_CoRR_Ba} proposed Layer
 Normalization, which performs normalization on a single example
using the mean and variance of the activations from the same
layer. Batch Normalization and Layer Normalization were later unified
by Ren \etal under the Division Normalization
framework~\cite{2017_ICLR_Ren}.  Other attempts to improve Batch
Normalization for small batch sizes, include Batch Renormalization
\cite{2017_NIPS_Ioffe} and Stream
Normalization~\cite{2016_axive_Liao}. There have also been efforts to
adapt Batch Normalization to Recurrent Neural
Networks~\cite{2016_ICASSP_Laurent,2016_CoRR_Cooijmans}. Our work
extends Batch Normalization by decorrelating the activations, which
is a direction orthogonal to all these prior works.

Our work is closely related to Natural Neural
Networks~\cite{2015_NIPS_Desjardins,2017_ICML_Luo}, which whiten activations by
periodically estimating and updating a whitening matrix. Our work
differs from Natural Neural Networks in two important ways. First,
Natural Neural Networks perform whitening by treating the mean and the
whitening matrix as model parameters as opposed to functions of the
input examples, which, as pointed out by Ioffe \&
Szegedy~\cite{2015_ICML_Ioffe,2017_NIPS_Ioffe}, can cause instability
in training, with symptoms such as divergence or gradient
explosion. Second, during training, a Natural Neural Network uses a
running estimate of the mean and whitening matrix to perform whitening
for each mini-batch; as a result, it cannot ensure that the transformed
activations within each batch are in fact whitened, whereas in our
case the activations with a mini-batch are guaranteed to be
whitened. Natural Neural Networks thus may suffer instability in
training very deep neural networks~\cite{2015_NIPS_Desjardins}.

Another way to obtain decorrelated activations is to introduce
additional regularization in the loss
function~\cite{2016_ICLR_Cogswell,2016_ICDM_Xiong}. Cogswell \etal
\cite{2016_ICLR_Cogswell} introduced the DeCov loss on the activations
as a regularizer to encourage non-redundant representations.  Xiong
\etal \cite{2016_ICDM_Xiong} extends \cite{2016_ICLR_Cogswell} to
learn group-wise decorrelated representations. Note that these
methods are not designed for speeding up training. In fact,
empirically they often slow down training \cite{2016_ICLR_Cogswell},
probably because decorrelated activations are part of the learning
objective and thus may not be achieved until later in training.

Our approach is also related to work that implicitly normalizes
activations by either normalizing the network weights---\eg through
re-parameterization
techniques~\cite{2016_CoRR_Salimans,Huang2017ICCV,2017_Huang_OWN},
Riemannian optimization methods \cite{2017_Huang_PBWN,2017_Corr_Cho},
or additional weight regularization
~\cite{1992_WD_Krogh,2015_NIPS_Neyshabur,2017_ICLR_Pau}---or by
designing special scaling coefficients and bias values that can induce
normalized activations under certain
assumptions~\cite{2016_ICML_Arpit}.  Our work bears some similarity to
that of Huang \etal~\cite{2017_Huang_OWN}, which also back-propagates
gradients through a ZCA-like normalization transform that involves
eigen-decomposition. But the work by Huang \etal normalizes weights
instead of activations, which leads to significantly different
derivations especially with regards to convolutional layers; in
addition, unlike ours, it does not involve a separately estimated
whitening matrix during inference, nor does it discuss the stochastic
axis swapping issue. Finally, all of these works including
~\cite{2017_Huang_OWN} are orthogonal to ours in the sense that their
normalization is data independent, whereas ours is data dependent. In
fact, as shown in
\cite{Huang2017ICCV,2017_Corr_Cho,2017_Huang_OWN,2017_Huang_PBWN},
data-dependent and data-independent normalization can be combined to
achieve even greater improvement.

\section{Decorrelated Batch Normalization}

Let $\mathbf{X} \in \mathbb{R}^{d\times m}$ be a data matrix that represents inputs to a
layer in a mini-batch of size $m$.
Let $\mathbf{x}_i \in \mathbb{R}^d$ be the $i$-th column vector of
$\mathbf{X}$, \ie the $d$-dimensional input from the $i$-th example.
The whitening transformation $\mathit{\phi}: \mathbb{R}^{d\times m}
\rightarrow \mathbb{R}^{d\times m}$ is defined as
\begin{equation} \phi(\mathbf{X})
= \Sigma^{-1/2} (\mathbf{X} - \mathbf{\mu} \cdot \mathbf{1}^T),
\label{eqn:whiten}
\end{equation}
where $\mathbf{\mu} = \frac{1}{m} \mathbf{X} \cdot \mathbf{1}$ is the
mean of $\mathbf{X}$, $\Sigma = \frac{1}{m} (\mathbf{X} - \mathbf{\mu}
\cdot \mathbf{1}^T) (\mathbf{X} - \mathbf{\mu} \cdot \mathbf{1}^T)^T +
\epsilon \mathbf{I}$ is the covariance matrix of the centered
$\mathbf{X}$, $\mathbf{1}$ is a column vector of all ones, and
$\epsilon>0$ is a small positive number for numerical stability
(preventing a singular $\Sigma$).  The whitening transformation
$\phi(\mathbf{X})$ ensures that for the transformed data
$\widehat{\mathbf{X}}=\phi(\mathbf{X})$ is whitened, \ie,
$\widehat{\mathbf{X}} \widehat{\mathbf{X}}^T=\mathbf{I}$.

Although Eqn.~\ref{eqn:whiten} gives an analytical form of the
whitening transformation, this transformation is in fact not
unique. The reason is that $\Sigma^{-1/2}$, the inverse square root of the
covariance matrix, is defined only up to rotation, and as a result
there exist infinitely many whitening transformations.
Thus, a natural question is whether the specific choice of
$\Sigma^{-1/2}$ matters, and if so, which choice to use.

To answer this question, we first discuss a phenomenon we call \emph{stochastic
  axis swapping} and show that not all whitening transformations are equally desirable.

\subsection{Stochastic Axis Swapping}
\label{para_3_1}
Given a data point represented as a vector $\mathbf{x} \in
\mathbb{R}^{d}$ under the standard basis, its representation under
another orthogonal basis $\{\mathbf{d}_1, ..., \mathbf{d}_d \}$ is
$\hat{\mathbf{x}}=\mathbf{D}^T \mathbf{x}$, where
$\mathbf{D}=[\mathbf{d}_1, ..., \mathbf{d}_d]$ is an orthogonal
matrix. We define \emph{stochastic axis swapping} as follows:
 \begin{defn}
 \label{def1} Assume a training algorithm that iteratively update weights using a batch of
 randomly sampled data points per iteration. \textbf{Stochastic axis swapping} occurs when a data point
 $\mathbf{x}$ is transformed to be $\hat{\mathbf{x}}_{1}=\mathbf{D}_1^T \mathbf{x}$ in one
 iteration and $\hat{\mathbf{x}}_2=\mathbf{D}_2^T
\mathbf{x}$ in another iteration such that $\mathbf{D}_1=\mathbf{P} \mathbf{D}_2 $
where $\mathbf{P} \neq \mathbf{I}$ is a permutation matrix solely determined by the statistics of
a batch.
 \end{defn}

\emph{Stochastic axis swapping} makes training difficult, because
the random permutation of the input dimensions can greatly
confuse the learning algorithm---in the extreme case where the permutation is completely
random, what remains is only a bag of activation values (similar to
scrambling all pixels in an image), potentially resulting in an extreme loss of information and discriminative power.

Here, we demonstrate that the whitening of activations, if not done
properly, can cause \emph{stochastic axis swapping} in training neural
networks.  We start with standard PCA
whitening~\cite{2015_NIPS_Guillaume}, which computes $\Sigma^{-1/2}$
through eigen decomposition: $\Sigma^{-1/2}_{pca} = \Lambda^{-1/2}
\mathbf{D}^T$,
where $\Lambda=\mbox{diag}(\sigma_1, \ldots,\sigma_d)$ and $\mathbf{D}=[\mathbf{d}_1, ...,
\mathbf{d}_d]$ are the eigenvalues and eigenvectors of $\Sigma$, \ie $\Sigma = \mathbf{D}
\Lambda \mathbf{D}^T$.  That is, the original data point (after centering) is rotated by
$\mathbf{D}^T$ and then scaled by $\Lambda^{-1/2}$. Without loss of generalization, we assume that $\mathbf{d}_i$ is
unique by fixing the sign of its first element. A first opportunity for stochastic
axis swapping is that the columns (or rows) of $\Lambda$ and $\mathbf{D}$ can be permuted
while still giving a valid whitening transformation. But this is easy to fix---we can
commit to a unique $\Lambda$ and $\mathbf{D}$ by ordering the eigenvalues non-increasingly.

\begin{figure}[t]
  \centering
\subfigure[]{
\begin{minipage}[c]{.46\linewidth}
\centering
  \includegraphics[width=3.2cm]{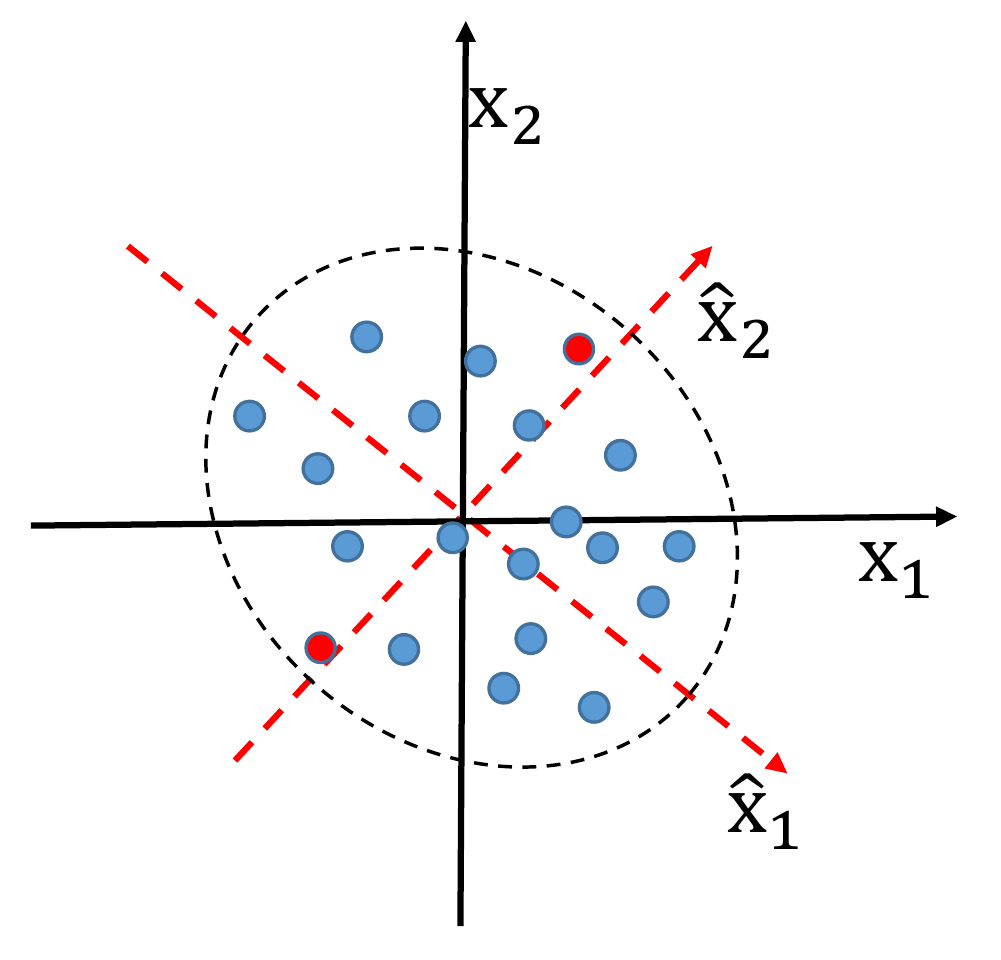}
\end{minipage}
}%
\subfigure[]{
\begin{minipage}[c]{.46\linewidth}
\centering
  \includegraphics[width=3.2cm]{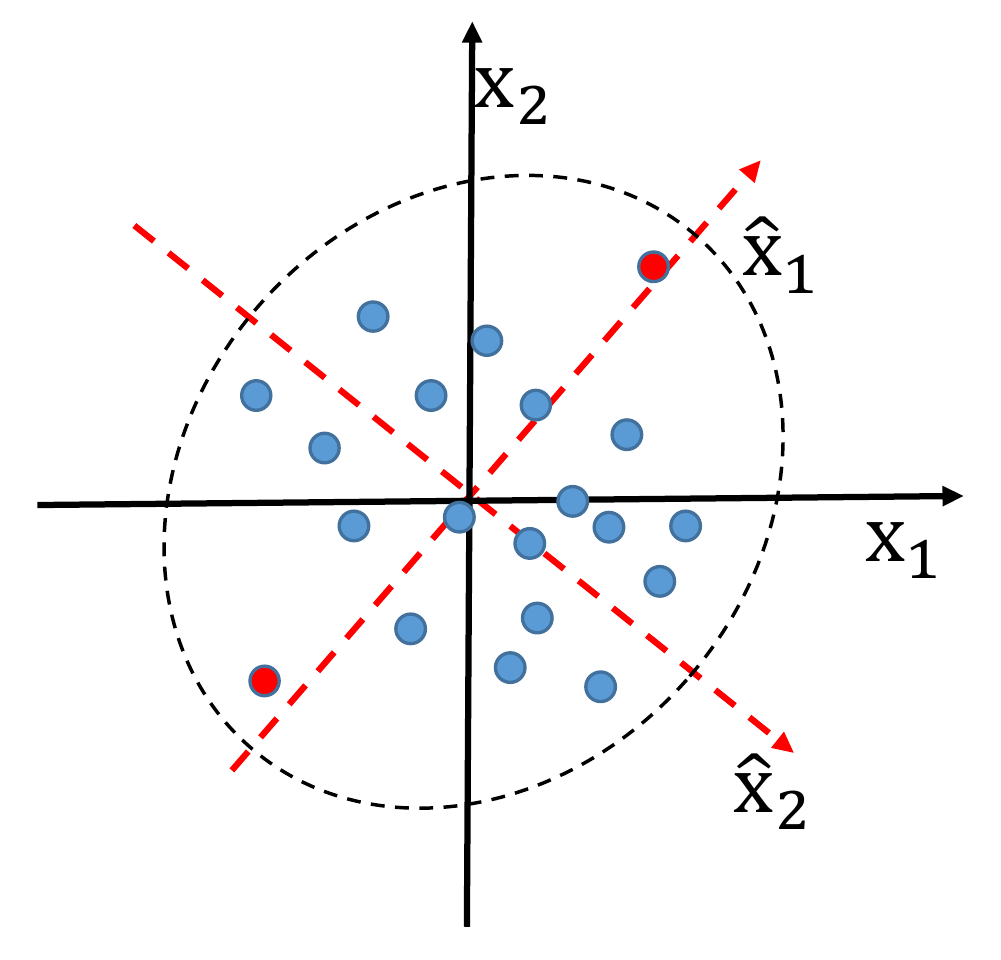}
\end{minipage}
}
  \caption{Illustration that PCA whitening suffers from stochastic axis swapping. (a)
    The axis alignment of PCA whitening in the initial iteration; (b) The axis alignment
    in another iteration. }
  \label{fig:idea}
\end{figure}

But it turns out that ensuring a unique $\Lambda$ and $\mathbf{D}$ is insufficient to
avoid stochastic axis swapping. Fig.~\ref{fig:idea} illustrates
an example. Given a mini-batch of data points in one iteration as shown in Fig.~\ref{fig:idea}(a), PCA whitening rotates them by
$\mathbf{D}^T=[\mathbf{d}_1^T, \mathbf{d}_2^T]^T$ and stretches them along the new axis system by
$\Lambda^{-1/2}=diag(1/\sqrt{\sigma_1}, 1/\sqrt{\sigma_2})$, where
$\sigma_1>\sigma_2$. Considering another iteration shown in Figure \ref{fig:idea}(b),
where all data points except the red points are the same, it has the same
eigenvectors with different eigenvalues, where $\sigma_1<\sigma_2$.
In this case, the new rotation matrix is
$(\mathbf{D}^{'})^T=[\mathbf{d}_2^T, \mathbf{d}_1^T]^T$ because we always order the eigenvalues
non-increasingly. The blue data points thus have two different representations with
the axes swapped.

\begin{figure}[t]
  \centering
\subfigure[2 layer MLP]{
\begin{minipage}[c]{.46\linewidth}
\centering
  \includegraphics[width=4.0cm]{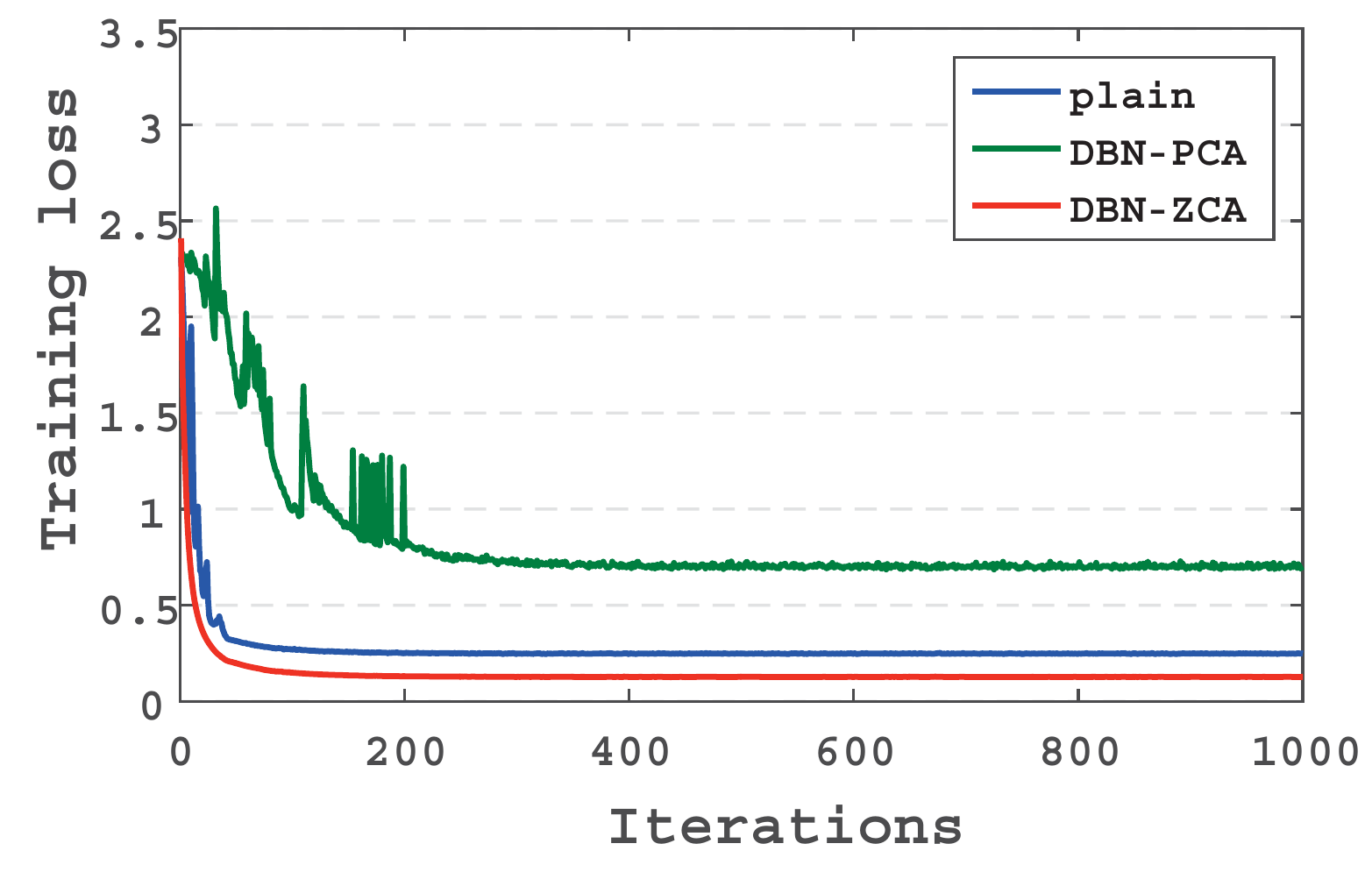}
\end{minipage}
}%
\subfigure[4 layer MLP]{
\begin{minipage}[c]{.46\linewidth}
\centering
  \includegraphics[width=4.0cm]{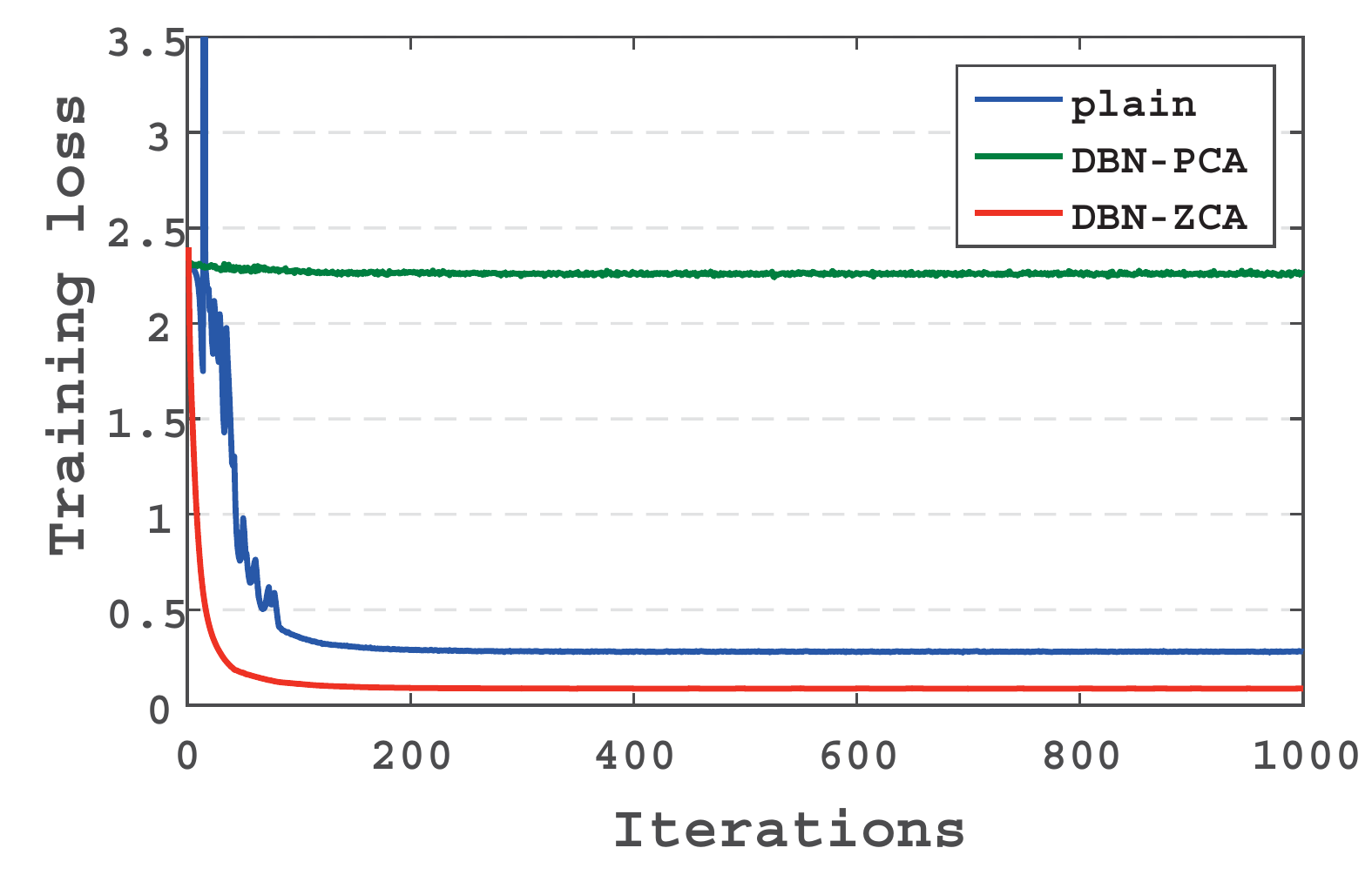}
\end{minipage}
}

\caption{\small Illustration of different whitening methods in
  training an MLP on MNIST. We use full batch gradient descent and report
  the best results with respect to the training loss among learning
  rates=$\{0.1, 0.5, 1, 5\}$.  (a) and (b) show the training loss of the
  2-layer and 4-layer MLP, respectively. The number of neurons in each
  hidden layer is 100.  We refer to the network without whitening
  activation as `plain`, with PCA whitening activation as DBN-PCA, and
  with ZCA whitening as DBN-ZCA.}
\label{fig:exp_MNIST_PCA}
\vskip -0.16in
\end{figure}

To further justify our conjecture, we perform an experiment on multilayer perceptrons
(MLPs) over the MNIST dataset as shown in Figure \ref{fig:exp_MNIST_PCA}. We refer to the
network without whitening activations as `plain' and the network with PCA whitening as
DBN-PCA.  We find that DBN-PCA has significantly inferior performance to
`plain'. Particularly, on the 4-layer MLP, DBN-PCA behaves similarly to random guessing, which
implies that it causes severe \emph{stochastic axis swapping}.

The stochastic axis swapping caused by PCA whitening exists because
the rotation operation is executed over varied activations. Such
variation is a result of two factors: (1) the activations can change
due to weight updates during training, following the \emph{internal
  covariate shift} described in~\cite{2015_ICML_Ioffe}; (2) the
optimization is based on random mini-batches, which means that each
batch will contain a different random set of examples in each training
epoch.

A similar phenomenon is also observed in \cite{2017_Huang_OWN}. In
this work, PCA-style orthogonalization failed to learn orthogonal
filters effectively in neural networks. However, no further analysis
was provided to explain why this is the case.

\subsection{ZCA Whitening}

 To address the stochastic axis swapping problem, one straightforward idea is to rotate
 the transformed input back using the same rotation matrix $\mathbf{D}$:
 {\setlength\abovedisplayskip{6pt}
\setlength\belowdisplayskip{6pt}
\begin{equation}
\Sigma^{-1/2} = \mathbf{D}\Lambda^{-1/2} \mathbf{D}^T.
\end{equation}
}
\hspace{-0.05in}
In other words, we scale along the eigenvectors to
get the whitened activations under the original axis system. Such
whitening is known as ZCA whitening \cite{1997_Vr_Bell}, and has been
shown to minimize the distortion introduced by whitening under L2
distance \cite{1997_Vr_Bell,2017_AS_Kessy,2017_Huang_OWN}.  We perform
the same experiments with ZCA whitening as we did with PCA whitening:
with MLPs on MNIST.  Shown in Figure \ref{fig:exp_MNIST_PCA}, ZCA
whitening (referred to as DBN-ZCA) improves training performance
significantly compared to no whitening (`plain') and DBN-PCA. This
shows that ZCA whitening is critical to addressing the stochastic axis
swapping problem.

\paragraph{Back-propagation}
It is important to note that the back-propagation through ZCA
whitening is non-trivial. In our DBN, the mean $\mathbf{\mu}$ and the
covariance $\Sigma$ are not parameters of the whitening transform
$\phi$, but are functions of the mini-batch data $\mathbf{X}$. We need
to back-propagate the gradients through $\phi$ as in
\cite{2015_ICML_Ioffe,2017_Huang_OWN}. Here, we use the results from
\cite{2015_ICCV_Ionescu} to derive the back-propagation formulations
of whitening: {\setlength\abovedisplayskip{6pt}
  \setlength\belowdisplayskip{6pt}
\begin{eqnarray}
\label{eqn:eig_backprop}
\frac{\partial L}{\partial \Sigma}&=\mathbf{D}\{ (\mathbf{K}^T \odot (\mathbf{D}^T \frac{\partial L}{\partial \mathbf{D}} )) +
(\frac{\partial L}{\partial \Lambda})_{diag}\} \mathbf{D}^T,
\end{eqnarray}
}
\hspace{-0.05in}where  $L$ is the loss function, $\mathbf{K} \in \mathbb{R}^{d\times d}$ is 0-diagonal with
$\mathbf{K}_{ij}= \frac{1}{\sigma_i  -\sigma_j} [i\neq j]$,
the $\odot$ operator is element-wise matrix multiplication, and $(\frac{\partial
  L}{\partial \Lambda})_{diag}$ sets the off-diagonal elements of $\frac{\partial
  L}{\partial \Lambda}$ as zero.

Detailed derivation can be found in the appendix. Here we only show the simplified formulation:
 {\setlength\abovedisplayskip{6pt}
\setlength\belowdisplayskip{6pt}
 \begin{eqnarray}
 \label{equ_backPro}
&\frac{\partial L}{\partial \mathbf{x}_i}&=
    \big( \frac{\partial L}{\partial \tilde{\mathbf{x}}_i} - \mathbf{f} + \tilde{\mathbf{x}}_i^T \mathbf{S}  -    \tilde{\mathbf{x}}_i^T \mathbf{M}  \big) \Lambda^{-1/2} \mathbf{D}^T,
\end{eqnarray}
}
\hspace{-0.05in}where $\mathbf{S}=2(\mathbf{K}^T \odot (\Lambda
\mathbf{F}_c^T +\Lambda^{\frac{1}{2}} \mathbf{F}_c
\Lambda^{\frac{1}{2}}))_{sym}$, $\mathbf{M}= (\mathbf{F}_c)_{diag}$,
$\mathbf{F}_c=\frac{1}{m} (\sum_{i=1}^{m} \frac{\partial L}{\partial
  \tilde{\mathbf{x}}_i}^T \tilde{\mathbf{x}}_i^T) $, and
$\mathbf{f}=\frac{1}{m} \sum_{i=1}^{m} \frac{\partial L}{\partial
  \tilde{\mathbf{x}}_i} $. The notation $(\cdot)_{sym}$ represents
symmetrizing the corresponding matrix.

\subsection{Training and Inference}
\label{sec:trainingAndInfer}
Decorrelated Batch Normalization (DBN) is a data-dependent whitening
transformation with back-propagation formulations. Like Batch
Normalization \cite{2015_ICML_Ioffe}, it can be inserted extensively
into a network.
Algorithms \ref{alg_forward} and \ref{alg_backprop} describe the
forward pass and the backward pass of our proposed DBN
respectively. During
training, the mean $\mathbf{\mu}$ and the whitening matrix
$\Sigma^{-1/2}$ are calculated within each mini-batch to ensure that
the activations are whitened for each mini-batch.
We also maintain the expected mean $\mathbf{\mu}_{E}$ and the expected
whitening matrix $\Sigma^{-1/2}_{E}$ for use during inference.
Specifically, during training, we initialize $\mathbf{\mu}_{E}$ as
$\mathbf{0}$ and $\Sigma^{-1/2}_{E}$ as $\mathbf{I}$ and update them
by running average as described in Line 10 and 11 of Algorithm
\ref{alg_forward}.

Normalizing the activations constrains the model's capacity for
representation. To remedy this, Ioffe and Szegedy
\cite{2015_ICML_Ioffe} introduce extra learnable parameters $\gamma$
and $\beta$ in Eqn.~\ref{bn_forward}. These learnable parameters often
marginally improve the performance in our observation. For DBN, we
also recommend to use learnable parameters.  Specifically, the
learnable parameters can be merged into the following ReLU activation
\cite{2010_ICML_Nair}, resulting in the Translated ReLU (TReLU)
\cite{2017_Corr_Xiang}.

For a convolutional neural network, the input to the DBN
transformation is $\mathbf{X}_C \in \mathbb{R}^{h \times w \times d
  \times m} $ where $h$ and $w$ indicate the height and width of
feature maps, and $d$ and $m$ are the numbers of feature maps and examples
respectively.
Following \cite{2015_ICML_Ioffe}, we view each spatial position of the
feature map as a sample. We thus unroll $\mathbf{X}_C$ as $\mathbf{X}
\in \mathbb{R}^{ d \times (m h w)}$ with $ m h w$ examples and $d$
feature maps. The whitening operation is performed over the unrolled
$\mathbf{X}$.

\begin{algorithm}[tb]
   \caption{Forward pass of DBN for each iteration.}
   \label{alg_forward}
\begin{algorithmic}[1]
\begin{small}
    \STATE \textbf{Input}: mini-batch inputs $\{ \mathbf{x}_i, i=1,2...,m \}$, expected mean $\mathbf{\mu}_{E}$  and expected projection matrix $\Sigma^{-1/2}_{E}$.
    \STATE \textbf{Hyperparameters}:  $\epsilon$, running average momentum $\lambda$.
     \STATE \textbf{Output}: the ZCA-whitened activations $\{ \hat{\mathbf{x}}_i, i=1,2...,m \}$.

    \STATE	calculate: $\mathbf{\mu} = \frac{1}{m} \sum_{j=1}^{m} \mathbf{x}_j$.
    \STATE	calculate: $\Sigma = \frac{1}{m} \sum_{j=1}^{m} (\mathbf{x}_j-\mathbf{\mu} )(\mathbf{x}_j-\mathbf{\mu} )^T + \epsilon \mathbf{I} $.
    \STATE	execute eigenvalue decomposition:  $\Sigma=\mathbf{D} \Lambda \mathbf{D}^T$.
    \STATE calculate  PCA-whitening matrix: $\mathbf{U}=  \Lambda^{-1/2} \mathbf{D}^T$.
    \STATE calculate  PCA-whitened activation : $\tilde{\mathbf{x}}_i =  \mathbf{U} (\mathbf{x}_i - \mathbf{\mu}) $.
    \STATE  calculate  ZCA-whitened output: $\hat{\mathbf{x}}_i = \mathbf{D} \tilde{\mathbf{x}}_i $.
     \STATE  update: $\mathbf{\mu}_{E} \leftarrow (1-\lambda) ~\mathbf{\mu}_{E} + \lambda ~\mathbf{\mu}$.

     \STATE update: $\Sigma^{-1/2}_E \leftarrow (1-\lambda) \Sigma^{-1/2}_{E} + \lambda \mathbf{D} \mathbf{U} $.
\end{small}
\end{algorithmic}
\end{algorithm}

\begin{algorithm}[tb]
   \caption{Backward pass of DBN for each iteration.}
   \label{alg_backprop}
\begin{algorithmic}[1]
\begin{small}
    \STATE \textbf{Input}: mini-batch  gradients respect to whitened outputs $\{ \frac{\partial L}{\partial \hat{\mathbf{x}}_i}, i=1,2...,m \}$.
    Other auxiliary data from respective forward pass: (1) eigenvalues; (2) $\tilde{\mathbf{x}}$; (3) $\mathbf{D}$.

     \STATE \textbf{Output}: the gradients respect to the inputs $\{ \frac{\partial L}{\partial \mathbf{x}_i}, i=1,2...,m \}$.
    \STATE calculate the gradients respect to $\tilde{\mathbf{x}}$: $\frac{\partial L}{\partial \tilde{\mathbf{x}}_i}	 =\frac{\partial L}{\partial \hat{\mathbf{x}}_i}     \mathbf{D}$.
    \STATE calculate $\mathbf{f}=\frac{1}{m} \sum_{i=1}^{m} \frac{\partial L}{\partial \tilde{\mathbf{x}}_i}^T $.
    \STATE calculate 0-diagonal $\mathbf{K}$ matrix by $\mathbf{K}_{ij}= \frac{1}{\sigma_i  -\sigma_j} [i\neq j]$.
    \STATE generate diagonal matrix $\Lambda$ from eigenvalues.
    \STATE  calculate $\mathbf{F}_c=\frac{1}{m} (\sum_{i=1}^{m} \frac{\partial L}{\partial \tilde{\mathbf{x}}_i}^T \tilde{\mathbf{x}}_i^T) $ and $\mathbf{M}= (\mathbf{F}_c)_{diag}$.
    \STATE calculate $\mathbf{S}=2(\mathbf{K}^T \odot   (\Lambda \mathbf{F}_c^T +\Lambda^{\frac{1}{2}} \mathbf{F}_c \Lambda^{\frac{1}{2}}))_{sym}$.
     \STATE  calculate $\frac{\partial L}{\partial \mathbf{x}_i}$ by formula \ref{equ_backPro}.
\end{small}
\end{algorithmic}
\end{algorithm}

\subsection{Group Whitening}
\label{sec:group}

As discussed in Section \ref{sec_intro}, it is necessary to control
the extent of whitening such that there are sufficient examples in a batch for estimating the
whitening matrix. To do this we use ``group whitening'', specifically, we divide the activations
along the feature dimension with size $d$ into smaller groups of size
$k_G$ ($k_G < d$) and perform whitening within each group. The extent
of whitening is controlled by the hyperparameter $k_G$.
In the case $k_G=1$, Decorrelated Batch Normalization reduces to
the original Batch Normalization.

In addition to controlling the extent of whitening, group whitening
reduces the computational complexity \cite{2017_Huang_OWN}. Full
whitening costs $O(d^2 \max(m,d))$ for a batch of size $m$. When using
group whitening, the cost is reduced to $O(\frac{d}{k_G}(k_G^2(\max(m,
k_G))))$. Typically, we choose $k_G< m$, therefore the cost of group
whitening is $O(mdk_G)$.

\subsection{Analysis and Discussion}
\label{sec:analyze_discuss}
DBN extends BN such that the activations are decorrelated over
mini-batch data. DBN thus inherits the beneficial properties of BN,
such as the ability to perform efficient training with large learning
rates and very deep networks. Here, we further highlight the benefits
of DBN over BN, in particular achieving better dynamical isometry
\cite{2013_CoRR_Saxe} and improved conditioning.

\paragraph{Approximate Dynamical Isometry} Saxe \etal \cite{2013_CoRR_Saxe} introduce
dynamical isometry---the desirable property that occurs when the
singular values of the product of Jacobians lie within a small range
around $1$. Enforcing this property, even approximately, is beneficial
to training because it preserves the gradient magnitudes during
back-propagation and alleviates the vanishing and exploding gradient
problems \cite{2013_CoRR_Saxe}.  Ioffe and Szegedy
\cite{2015_ICML_Ioffe} find that Batch Normalization achieves
approximate dynamical isometry under the assumption that (1) the
transformation between two consecutive layers is approximately linear,
and (2) the activations in each layer are Gaussian and
uncorrelated. Our DBN inherently satisfies the second assumption, and
therefore is more likely to achieve dynamical isometry than BN.

\paragraph{Improved Conditioning}
\cite{2015_NIPS_Desjardins} demonstrated that whitening activations
results in a block diagonal Fisher Information Matrix (FIM) for each
layer under certain assumptions \cite{2016_ICML_Grosse}.
Their experiments show that such a block diagonal structure in the FIM
can improve the conditioning. The proposed method in
\cite{2015_NIPS_Desjardins}, however, cannot whiten the activations
effectively, as shown in \cite{2017_ICML_Luo} and also discussed in
Section \ref{sec_relatedWork}. DBN, on the other hand, does this
directly. Therefore, we conjecture that DBN can further improve the
conditioning of the FIM, and we justify this experimentally in Section
\ref{exp:MLP}.

\section{Experiments}
We start with experiments to highlight the effectiveness of Decorrelated Batch
Normalization (DBN) in improving the conditioning and speeding up convergence on
multilayer perceptrons (MLP). We then conduct comprehensive experiments to compare DBN and
BN on convolutional neural networks (CNNs).  In the last section, we
apply our DBN to residual networks on CIFAR-10, CIFAR-100 ~\cite{2009_TR_Alex} and
ILSVRC-2012 to show its power to improve modern network architectures.
The code to reproduce the experiments is available at \textcolor[rgb]{0.33,0.33,1.00}{https://github.com/umich-vl/DecorrelatedBN}.

We focus on classification tasks and the loss function is the negative
log-likelihood: $- \log P(\mathbf{y}|\mathbf{x})$. Unless otherwise
stated, we use random weight initialization as described in
\cite{1998_NN_Yann} and ReLU activations \cite{2010_ICML_Nair}.

\begin{figure}[t]
\centering
\hspace{-0.03\linewidth}
\subfigure[]{
  \includegraphics[width=0.48\linewidth]{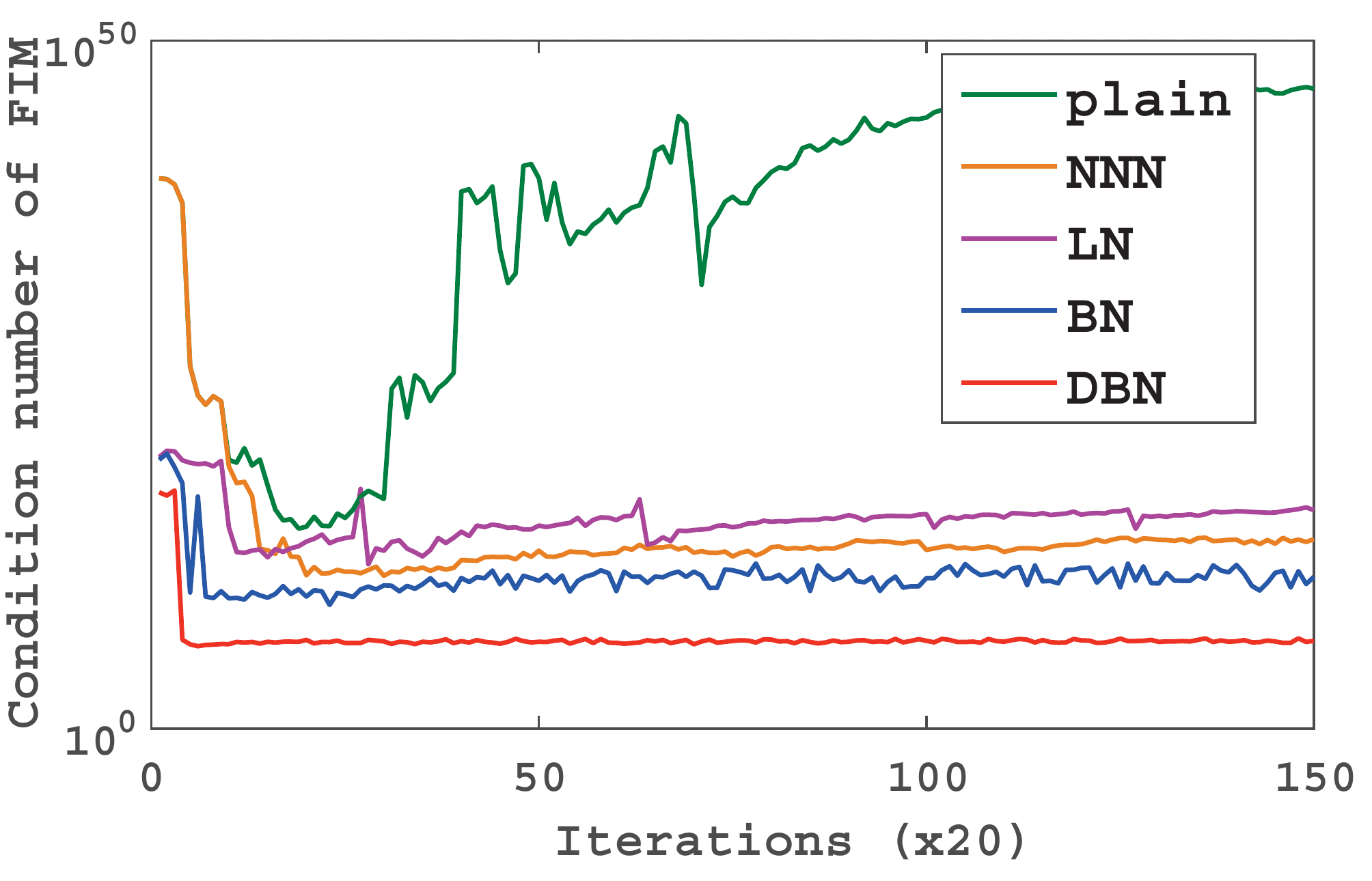}
  }
  \hspace{-0.03\linewidth}
\subfigure[]{
  \includegraphics[width=0.48\linewidth]{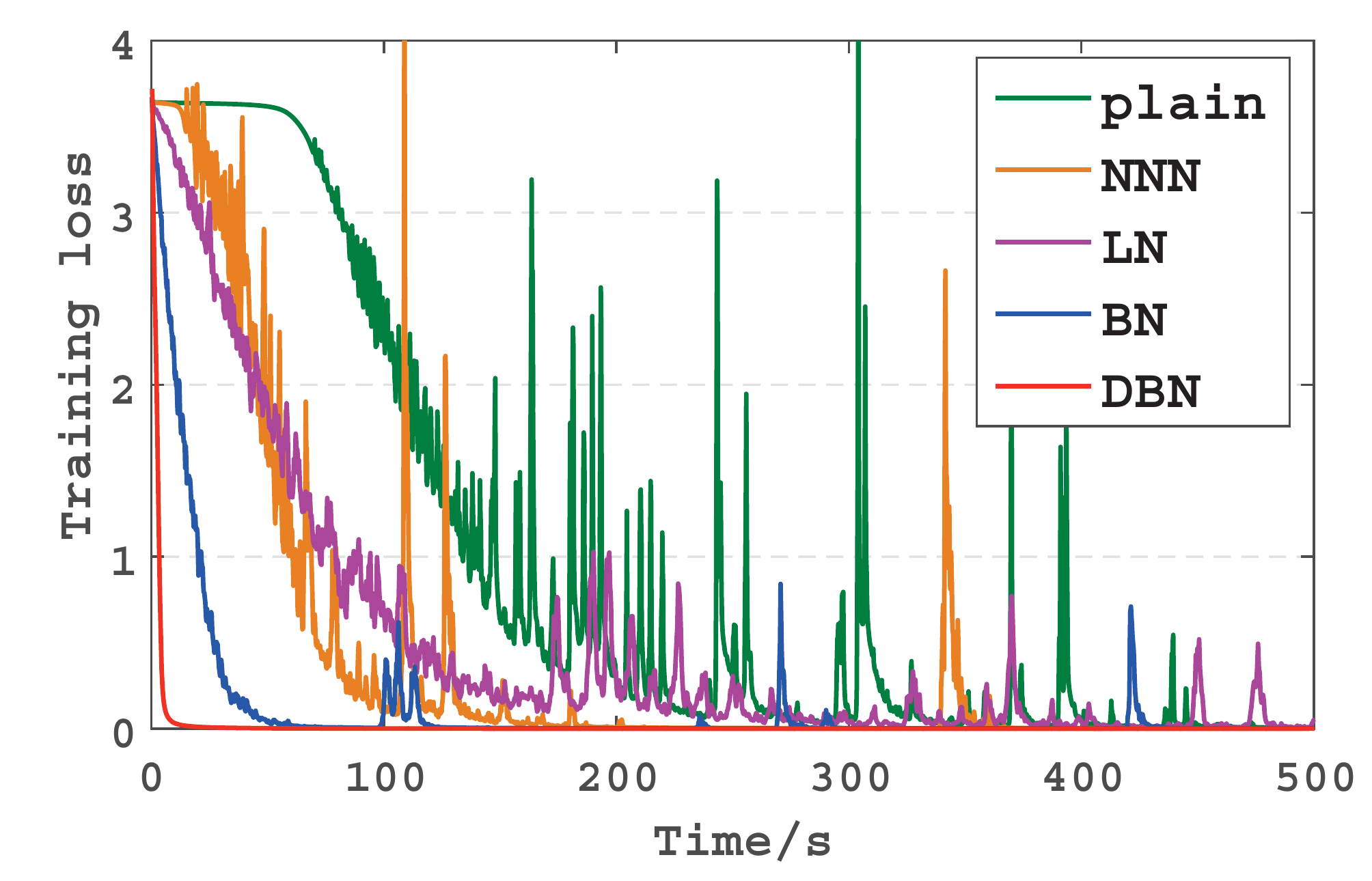}
  }
\caption{\small Conditioning analysis with MLPs trained on the Yale-B
  dataset. (a) Condition number (log-scale) of relative FIM as a
  function of updates in the last layer; (b) training loss with
  respect to wall clock time. }
\label{fig:exp_MLP_1}
\vskip -0.15in
\end{figure}

\begin{figure}[t]
\centering
\hspace{-0.03\linewidth}
\subfigure[]{
  \includegraphics[width=0.48\linewidth]{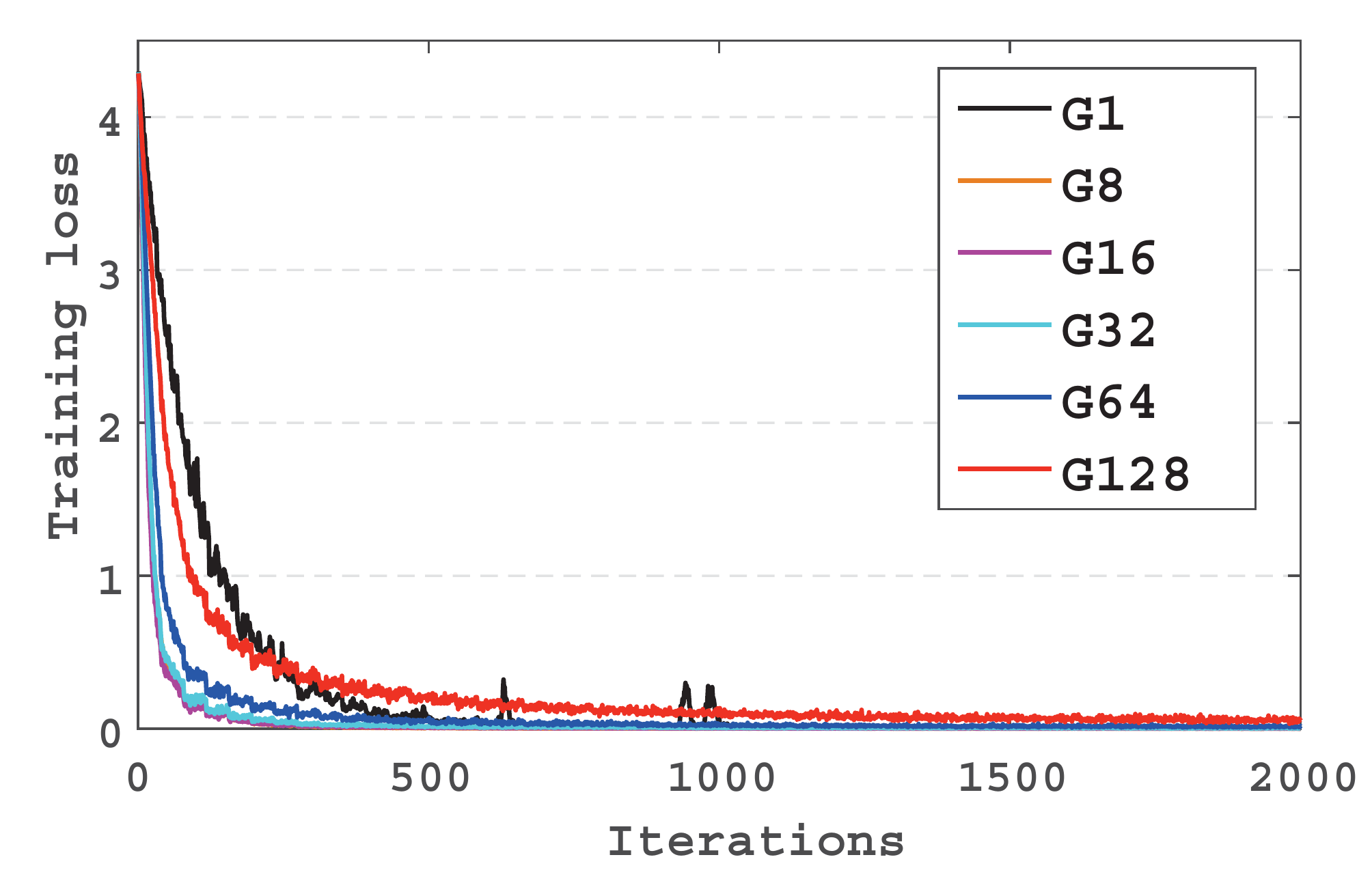}
  }
  \hspace{-0.03\linewidth}
\subfigure[]{
  \includegraphics[width=0.48\linewidth]{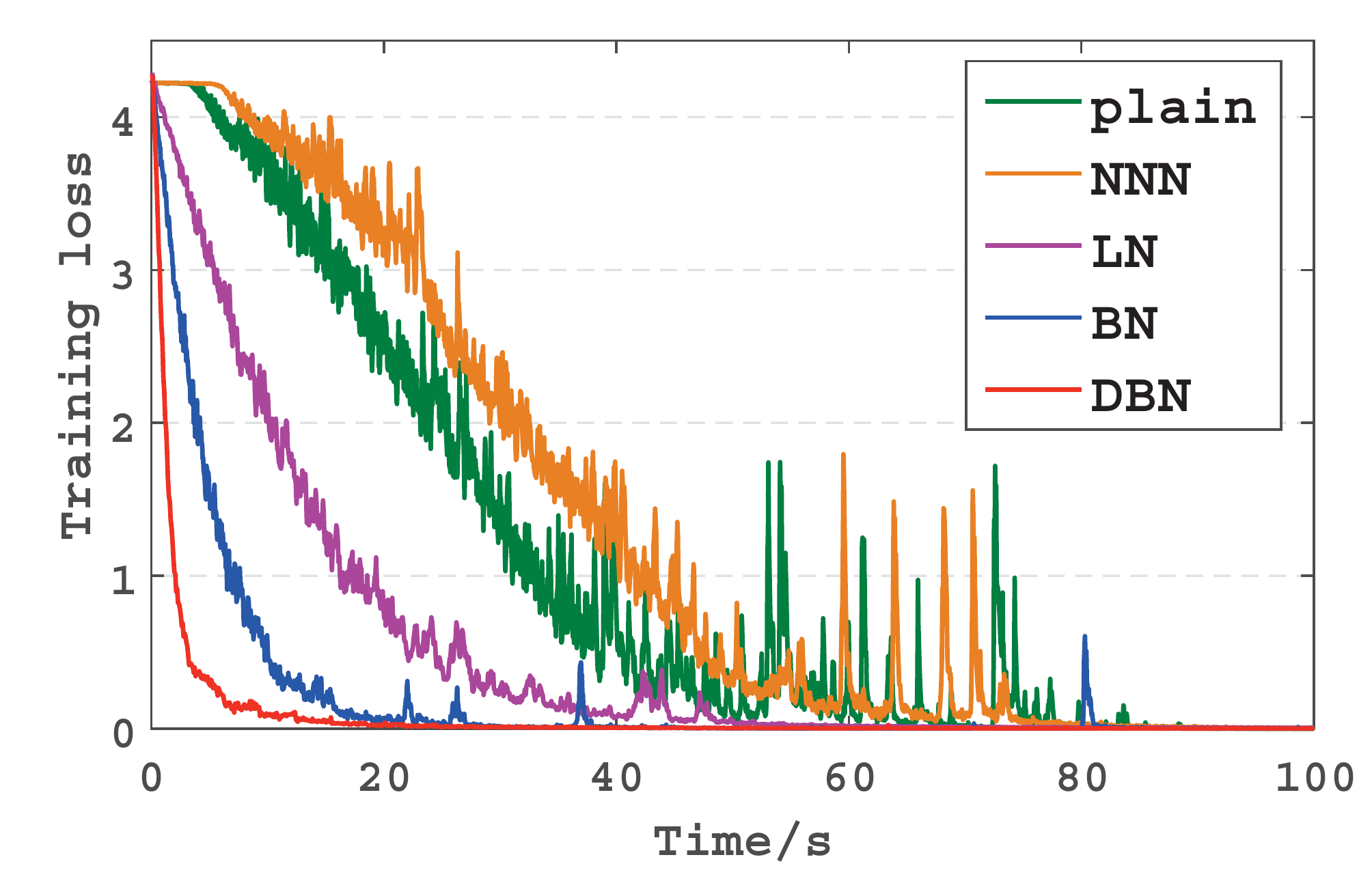}
  }

  \caption{\small  Experiments on MLP architecture over PIE dataset. (a) The effects of group size of DBN, where 'G$n$' indicates $k_G=n$; (b) Comparison of training loss with respect to wall clock time.}
  \label{fig:exp_MLP_2}
\end{figure}

\subsection{Ablation Studies on MLPs}
\label{exp:MLP}

In this section, we verify the effectiveness of our proposed method in
improving conditioning and speeding up convergence on MLPs. We also
discuss the effect of the group size on the tradeoff between the
performance and computation cost.  We compare against several
baselines, including the original network without any normalization
(referred to as `plain'), Natural Neural Networks (NNN)
\cite{2015_NIPS_Desjardins}, Layer Normalization (LN)
\cite{2016_CoRR_Ba}, and Batch Normalization (BN)
\cite{2015_ICML_Ioffe}.  All results are averaged over 5 runs.

\begin{figure*}[t]
\centering

\subfigure[Basic Configuration]{
  \includegraphics[width=0.23\linewidth]{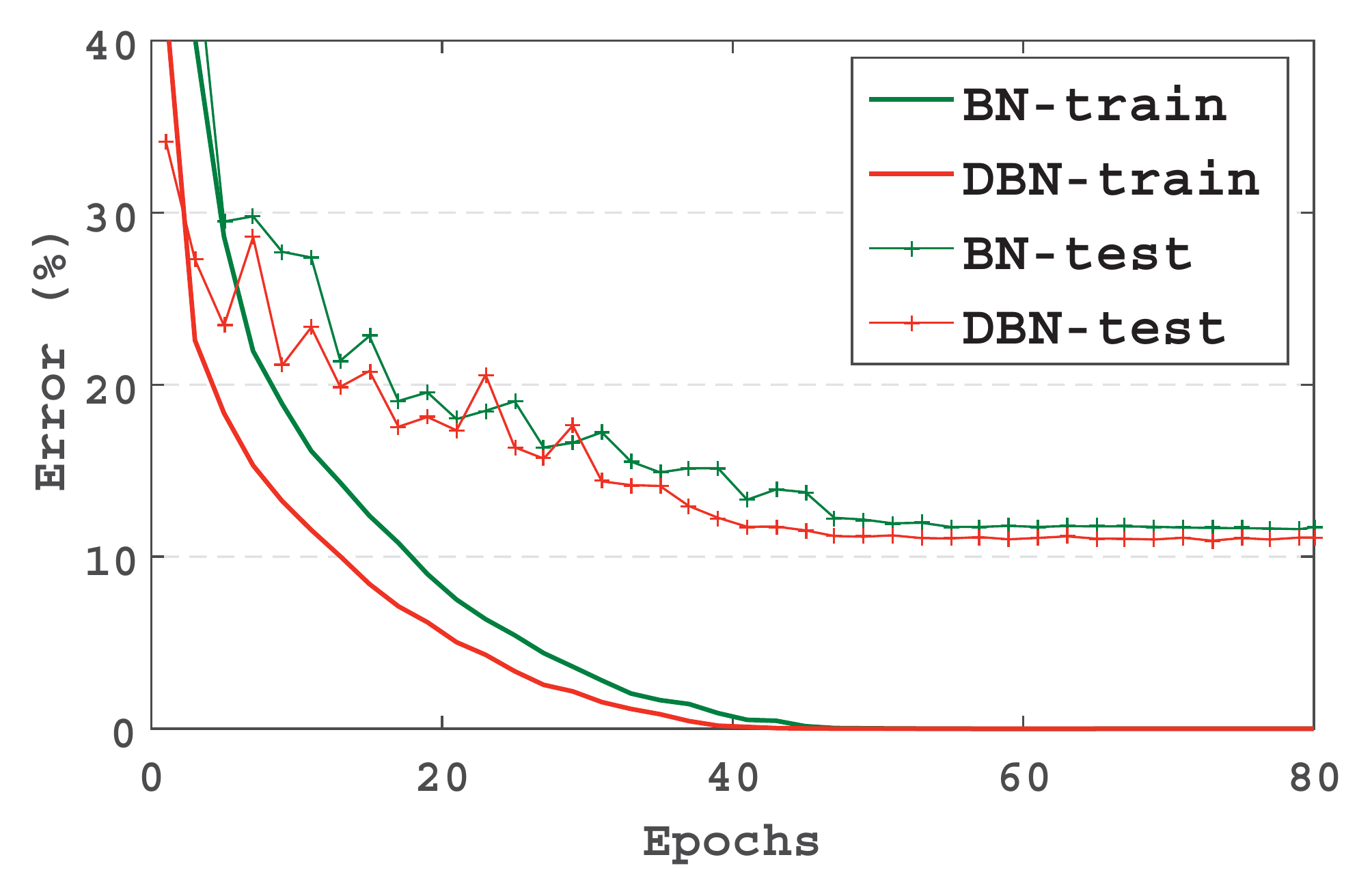}
  }
\subfigure[Adam optimization]{
  \includegraphics[width=0.23\linewidth]{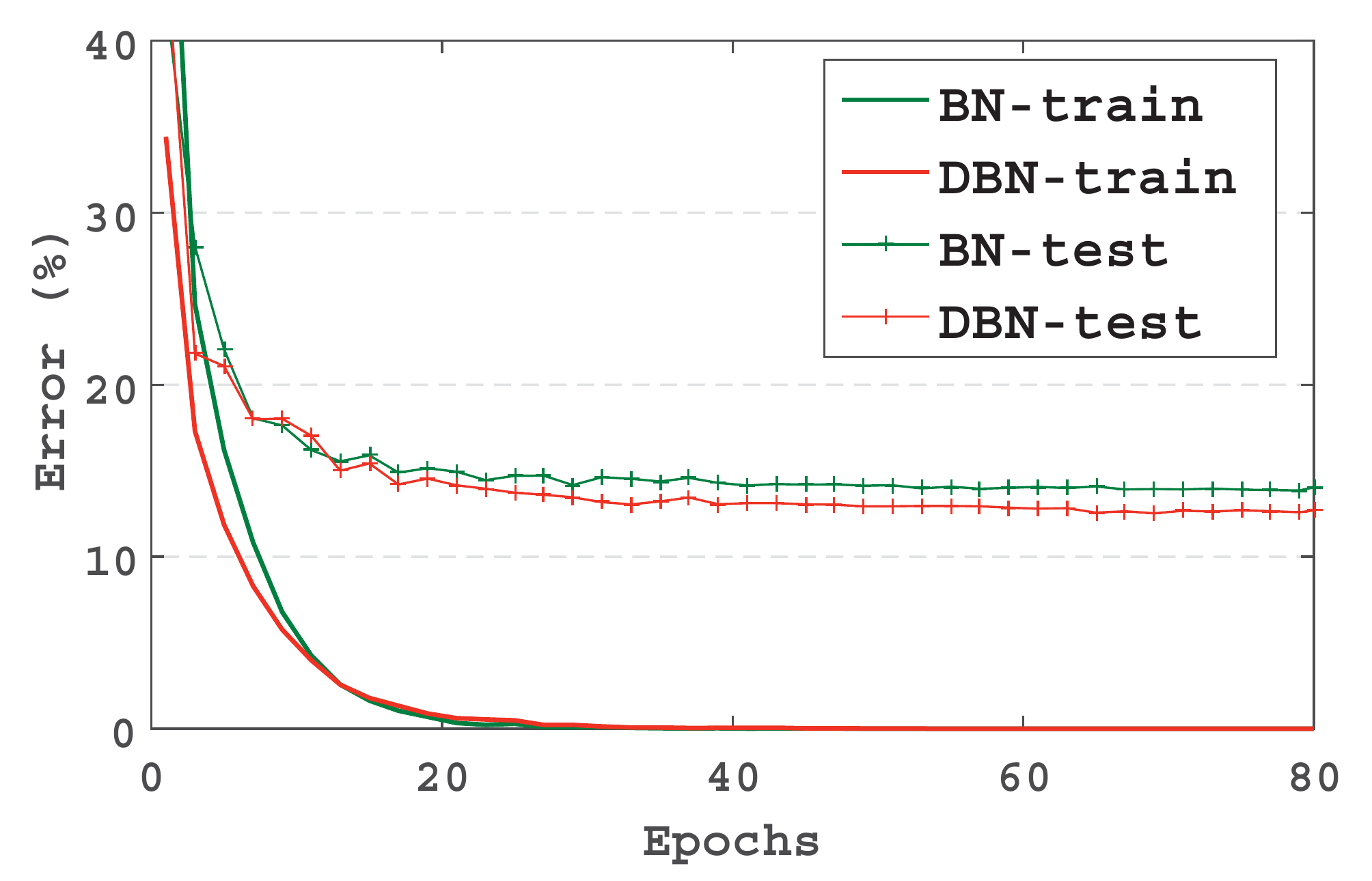}
  }
\subfigure[ELU non-linearity]{
  \includegraphics[width=0.23\linewidth]{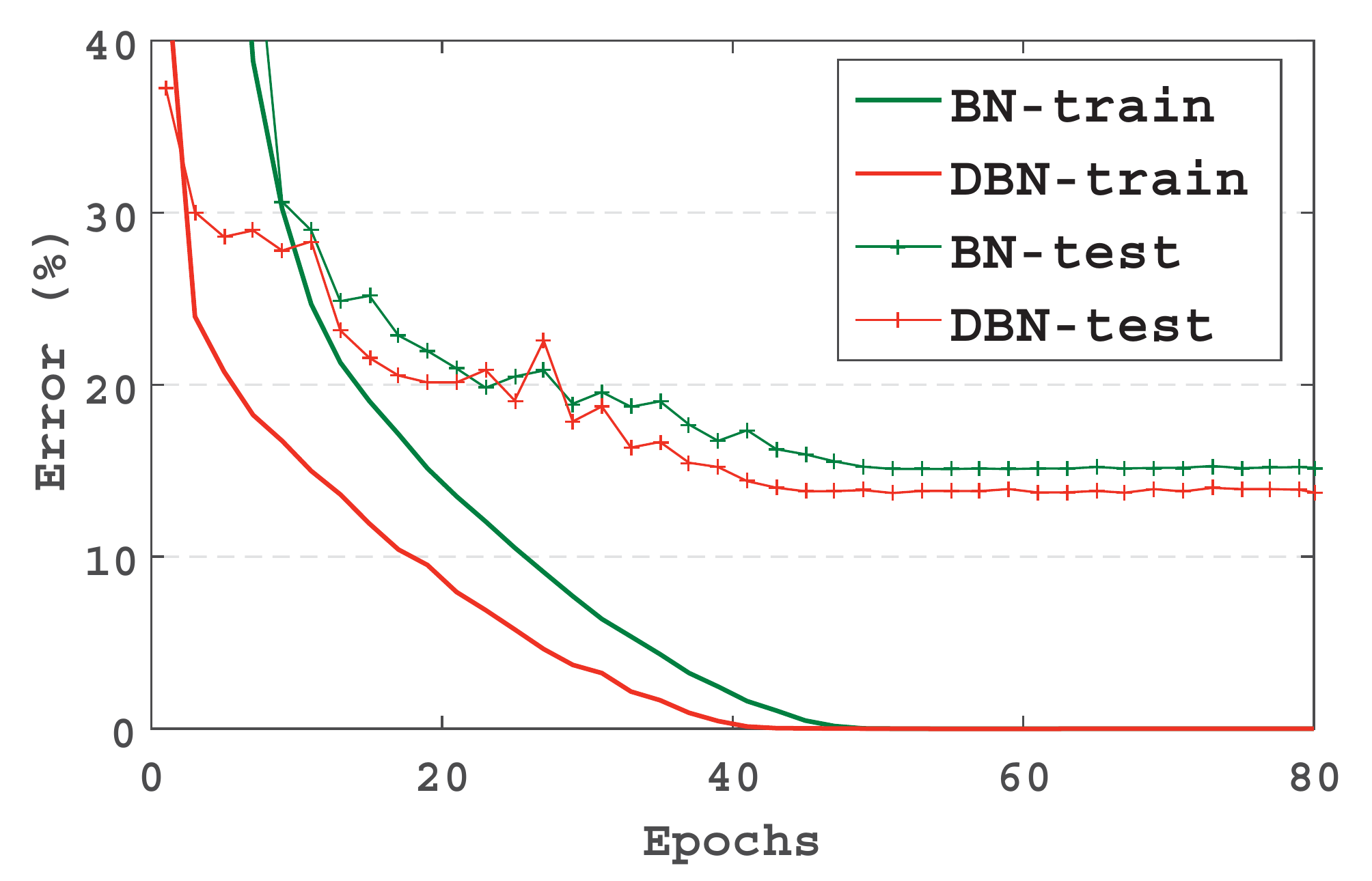}
  }
\subfigure[DBN/BN after non-linearity]{
  \includegraphics[width=0.23\linewidth]{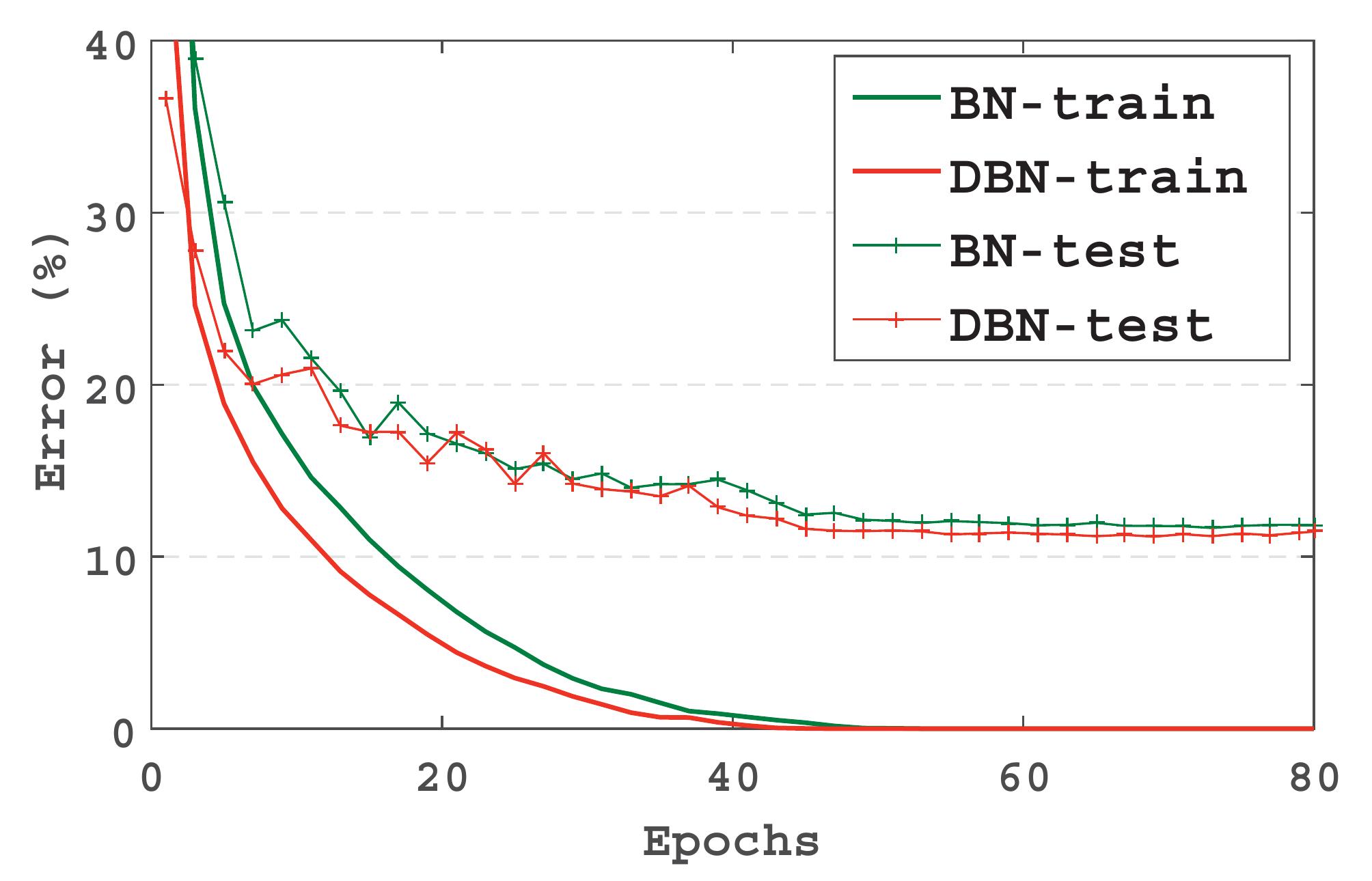}
  }

\caption{\small Comprehensive performance comparison between DBN and
  BN with the VGG-A architecture on CIFAR-10. We show the training
  accuracy (solid line) and test accuracy (line marked with plus) for
  each epoch.}
  \label{fig:exp_Vgg-A}
\end{figure*}

\paragraph{Conditioning Analysis}
We perform conditioning analysis on the Yale-B dataset
\cite{2001_TPAMI_YaleB}, specifically, the subset \cite{CHHH07} with
2,414 images and 38 classes. We resize the images to 32$\times$32 and
reshape them as 1024-dimensional vectors.  We then convert the images
to grayscale in the range $[0, 1]$ and subtract the per-pixel mean.

For each method, we train a 5-layer MLP with the numbers of neurons in
each hidden layer=$\{128,64,48,48\}$ and use full batch gradient
descent.  Hyper-parameters are selected by grid search based on the
training loss.  For all methods, the learning rate is chosen from $\{
0.1,0.2,0.5,1,2 \}$. For NNN, the revised term $\varepsilon$ is one of
$\{ 0.001,0.01,0.1,1 \}$ and the natural re-parameterization interval
$T$ is one of $\{20, 50, 100, 200, 500\}$.

We evaluate the condition number of the relative Fisher Information Matrix
(FIM)~\cite{2016_CoRR_Sun} with respect to the last layer.
Figure \ref{fig:exp_MLP_1} (a) shows the evolution of the condition
number over training iterations. Figure \ref{fig:exp_MLP_1} (b) shows
the training loss over the wall clock time. Note that the experiments
are performed on CPUs and the model with DBN is $2\times$ slower than
the model with BN per iteration. From both figures, we see that NNN,
LN, BN and DBN converge faster, and achieve better conditioning
compared to `plain'. This shows that normalization is able to make the
optimization problem easier. Also, DBN achieves the best conditioning
compared to other normalization methods, and speeds up convergence
significantly.

\paragraph{Effects of Group Size}
As discussed in Section \ref{sec:group}, the group size $k_G$ controls
the extent of whitening. Here we show the effects of the
hyperparameter $k_G$ on the performance of DBN.  We use a subset
\cite{CHHH07} of the PIE face recognition \cite{2002_FGR_Terence}
dataset with 68 classes with 11,554 images.  We adopt the same
pre-processing strategy as with Yale-B.

We trained a 6-layer MLP with the numbers of neurons in each hidden
layer=$\{128,128,128,128,128\}$.  We use Stochastic Gradient Descent
(SGD) with a batch size of 256.  Other configurations are chosen in
the same way as the previous experiment. Additionally, we explore
group sizes in $\{ 1, 8, 16, 32, 64, 128\}$ for DBN. Note that when
$k_G=1$, DBN is reduced to the original BN without the extra learnable
parameters.

Figure \ref{fig:exp_MLP_2} (a) shows the training loss of DBN with
different group sizes.  We find that the largest (G128) and smallest
group sizes (G1) both have noticeably slower convergence compared to
the ones with intermediate group sizes such as G16. These
results show that (1) decorrelating activations over a mini-batch can
improve optimization, and (2) controlling the extent of whitening is
necessary, as the estimate of the full whitening matrix might be poor
over mini-batch samples.  Also, the eigendecomposition with small
group sizes (\eg 16) is less computationally expensive. We thus
recommend using group whitening in training deep models.

We also compared DBN with group whitening ($k_G=16$) to other
baselines and the results are shown in Figure \ref{fig:exp_MLP_2}
(b). We find that DBN converges significantly faster than other
normalization methods.

\subsection{Experiments on CNNs}
\label{exp_CNN}
We design comprehensive experiments to evaluate the performance of DBN
with CNNs against BN, the state-of-the-art normalization technique.
For these experiments we use the CIFAR-10 dataset~\cite{2009_TR_Alex},
which contains 10 classes, 50k training images, and 10k test images.

\subsubsection{Comparison of DBN and BN}
We compare DBN to BN over different experimental configurations,
including the choice of optimization method, non-linearities, and the
position of DBN/BN in the network.  We adopt the VGG-A architecture
\cite{2014_CoRR_Simonyan} for all experiments, and pre-process the
data by subtracting the per-pixel mean and dividing by the variance.

We use SGD with a batchsize of 256, momentum of 0.9 and weight decay
of 0.0005. We decay the learning rate by half every $T$
iterations.  The hyper-parameters are chosen by grid search over a
random validation set of 5k examples taken from the training set. The
grid search includes the initial learning rate $lr=\{ 1,2,4,8 \}$ and
the decay interval $T=\{1000, 2000,4000,8000 \}$. We set the
group size of DBN as $k_G=16$.  Figure \ref{fig:exp_Vgg-A} (a)
compares the performance of BN and DBN under this configuration.

We also experiment with other configurations, including using (1) Adam
\cite{2014_CoRR_Kingma} as the optimization method, (2) replacing ReLU
with another widely used non-linearity called Exponential Linear Units
(ELU)~\cite{2015_CoRR_Clevert}, and (3) inserting BN/DBN after the
non-linearity. All the experimental setups are otherwise the same,
except that Adam \cite{2014_CoRR_Kingma} is used with an initial
learning rate in $\{ 0.001, 0.005, 0.01, 0.05 \}$. The respective
results are shown in Figure \ref{fig:exp_Vgg-A} (b), (c) and (d).

In all configurations, DBN converges faster with respect to the epochs and
generalizes better, compared to BN. Particularly, in the four
experiments above, DBN reduces the absolute test error by $0.61\%$,
$1.34\%$, $1.44\%$ and $0.38\%$ respectively.  The results demonstrate
that our Decorrelated Batch Normalization outperforms Batch
Normalization in terms of optimization quality and regularization
ability.

\begin{figure}[t]
\centering
\hspace{-0.07\linewidth}
\subfigure[Deeper Networks]{
  \includegraphics[width=0.46\linewidth]{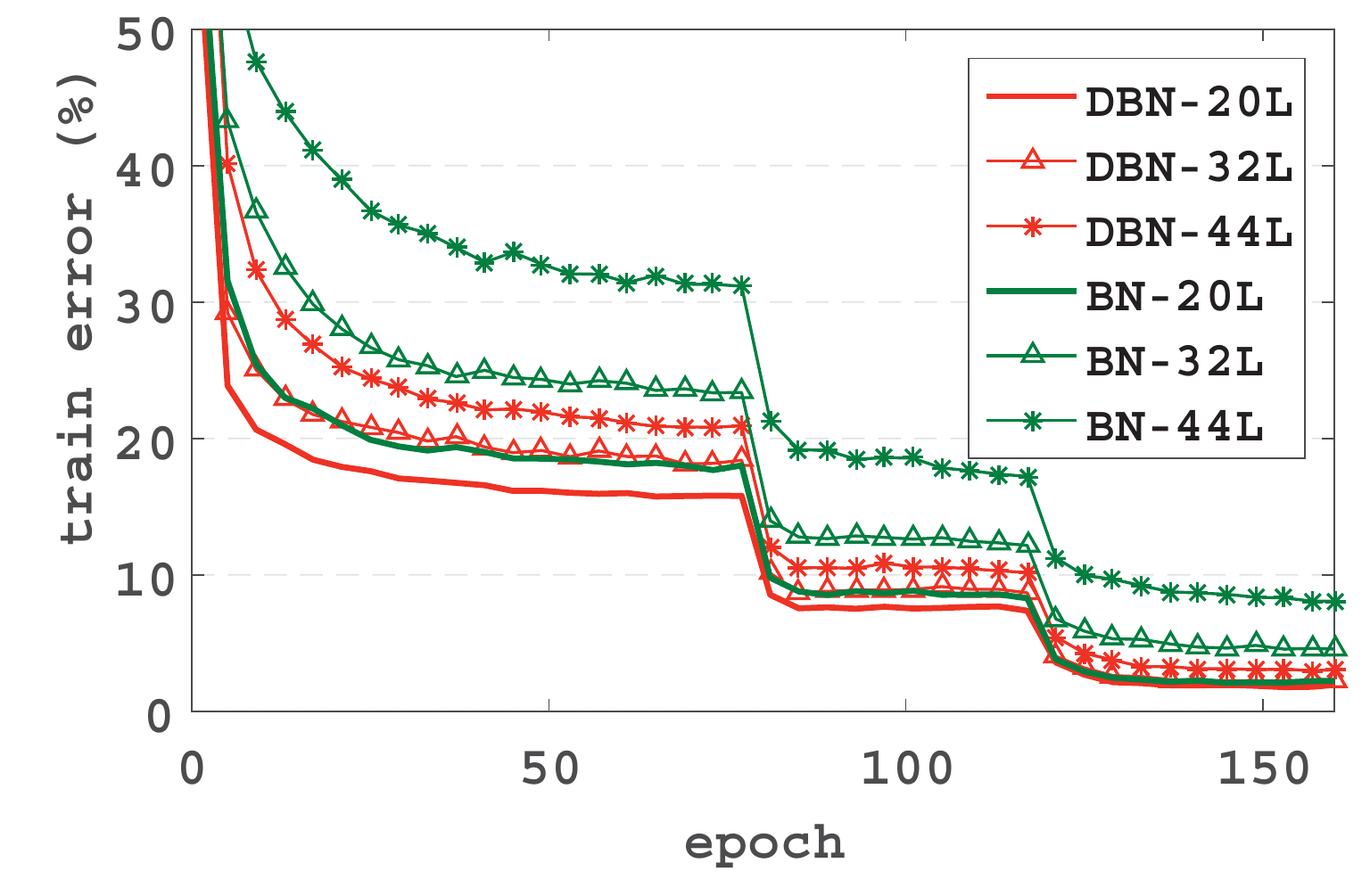}
  }
\hspace{-0.03\linewidth}
\subfigure[$4\times$ Learning Rate]{
  \includegraphics[width=0.46\linewidth]{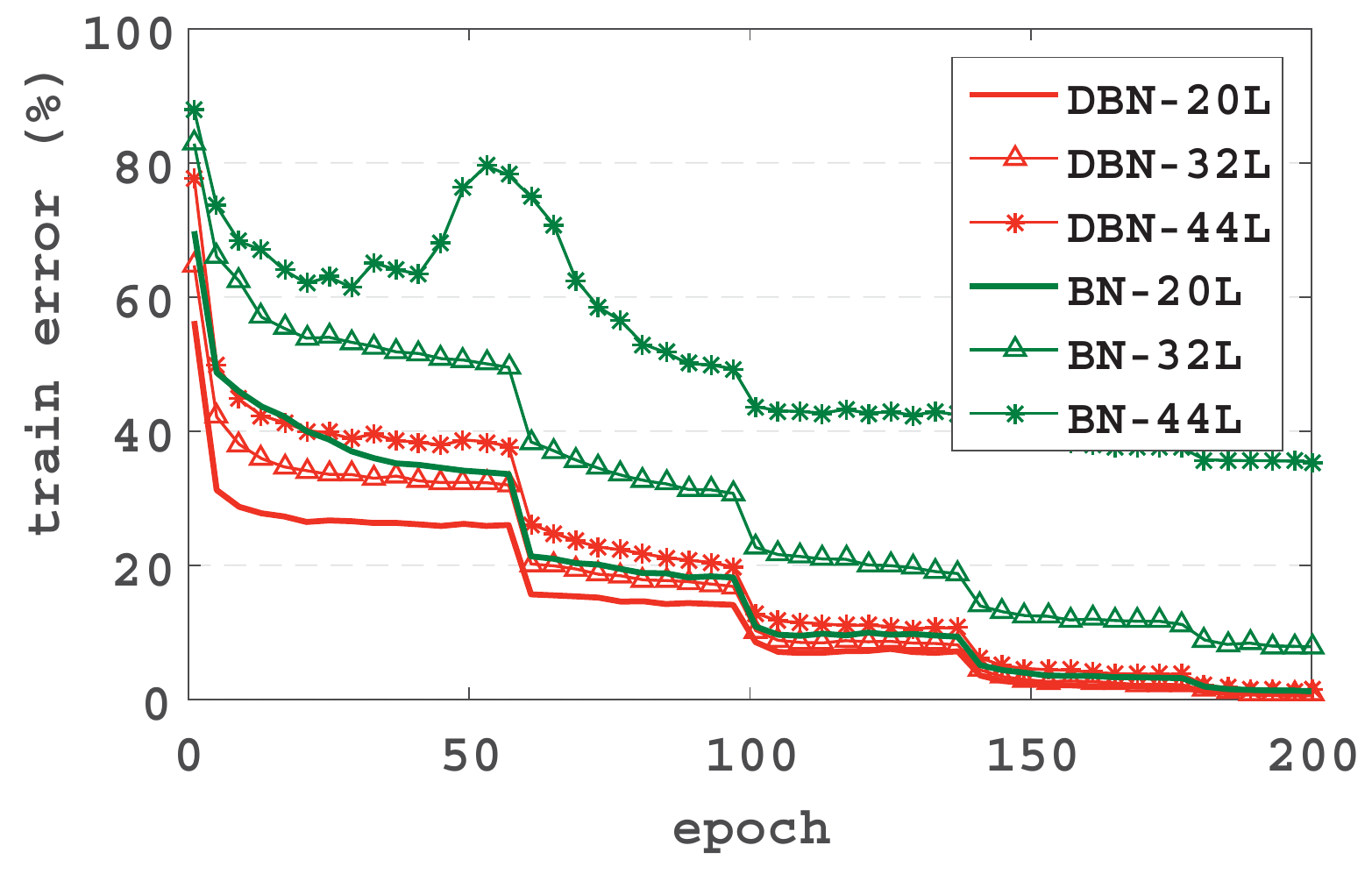}
  }
  \caption{\small DBN can make optimization easier and benefits from a
    higher learning rate. Results are reported on the \emph{S-plain}
    architecture over the CIFAR-10 dataset. (a) Comparison by varying
    the depth of network. '-\emph{n}L' means the network has \emph{n}
    layers; (b) Comparison by using a higher learning rate.}
  \label{fig:exp_deeper}
\end{figure}

\begin{table}[t]
\begin{center}
\begin{small}
\begin{tabular}{l|cccc}
\toprule
    Method & Res-20 & Res-32 & Res-44 & Res-56\\
\hline
Baseline*  &  8.75 &7.51 &7.17 & 6.97\\
Baseline  &  7.94 &7.31    &7.17  & 7.21  \\
DBN-L1 &  7.94 & 7.28 & 6.87 & 6.63 \\
DBN-scale-L1 &  \textbf{7.77}  &\textbf{6.94 } &\textbf{6.83 }   & \textbf{6.49}   \\
\bottomrule
\end{tabular}
\vskip 0.1in
\caption{Comparison of test errors ($\%$) with residual networks on
  CIFAR-10. `Res-$L$' indicates residual network with $L$ layers, and
  `Baseline*' indicates the results reported in \cite{2015_CVPR_He}
  with only one run. Our results are averaged over 5
  runs.} \label{table:resnet1}
\end{small}
\end{center}
\vskip -0.1in
\end{table}

\begin{table*}[]
  \centering
  \begin{small}
  \begin{tabular}{c|ccc|ccc}
    \toprule
    & \multicolumn{3}{c|}{CIFAR-10} &    \multicolumn{3}{c}{CIFAR-100}   \\
    Method & Baseline*~\cite{2016_CoRR_Zagoruyko}  & Baseline    & DBN-scale-L1  & Baseline*~\cite{2016_CoRR_Zagoruyko} & Baseline   & DBN-scale-L1 \\
  \hline
    WRN-28-10   & 3.89  & 3.99 $\pm$ 0.13 & \textbf{3.79  $\pm$ 0.09}   &  18.85 & 18.75 $\pm$ 0.28 & \textbf{18.36 $\pm$ 0.17} \\
    WRN-40-10    & 3.80  & 3.80 $\pm$ 0.11 & \textbf{3.74  $\pm$ 0.11}   & 18.3 & 18.7 $\pm$ 0.22 & \textbf{18.27 $\pm$ 0.19} \\
    \bottomrule
  \end{tabular}
    \vspace{0.1in}
\caption{Test errors ($\%$) on wide residual networks over CIFAR-10
  and CIFAR-100.  `Baseline' and `DBN-scale-L1' refer to the the
  results we perform, based on the released code of paper
  \cite{2016_CoRR_Zagoruyko}, and the results are shown in the format
  of `mean $\pm std$' computed over 5 random seeds.  `Baseline*'
  refers to the results reported by authors of
  ~\cite{2016_CoRR_Zagoruyko} on their Github. They report the median
  of 5 runs on WRN-28-10 and only perform one run on
  WRN-40-10.} \label{table:wr}

\end{small}
\end{table*}

\begin{table*}[htbp]
  \centering
  \begin{small}
  \begin{tabular}{c|cc|cc|cc|cc}
    \toprule
    & \multicolumn{2}{c|}{Res-18} &    \multicolumn{2}{c|}{Res-34} &\multicolumn{2}{c|}{Res-50} &\multicolumn{2}{c}{Res-101}   \\
    Method   & Top-1    & Top-5  & Top-1   & Top-5 & Top-1 & Top-5 & Top-1 & Top-5\\
  \hline
     Baseline* & -- & -- & -- & -- & 24.70  &  7.80   & 23.60 & 7.10  \\
     Baseline  & 30.21 & 10.87 & 26.53 & 8.59 & 24.87 &  7.58   & 22.54 & 6.38 \\
     DBN-scale-L1& \textbf{29.87} & \textbf{10.36} & \textbf{26.46} & \textbf{8.42} & \textbf{24.29}&\textbf{7.08}&\textbf{22.17} & \textbf{6.09}\\
    \bottomrule
  \end{tabular}
    \vspace{0.1in}
     \caption{Comparison of test errors ($\%$, single model and
     single-crop) on 18, 34, 50, and 101-layer residual networks
     on ILSVRC-2012.\hspace{\textwidth} `Baseline*' indicates that the results are obtained from the website: https://github.com/KaimingHe/deep-residual-networks}

     \label{table:ImageNet}
\end{small}
\end{table*}

\subsubsection{ Analyzing the Properties of DBN}
\label{exp_analyze}
We conduct experiments to support the conclusions from Section
\ref{sec:analyze_discuss}, specifically that DBN has better stability
and converges faster than BN with high learning rates in very deep
networks.  The experiments were conducted on the \emph{S-plain}
network, which follows the design of the residual network
\cite{2015_ICCV_He} but removes the identity maps and uses the same
feature maps for simplicity.

\paragraph{Going Deeper}
He \etal \cite{2015_ICCV_He} addressed the degradation problem for the
network without identity mappings: that is, when the network depth
increases, the training accuracy degrades rapidly, even when Batch
Normalization is used. In our experiments, we demonstrate that DBN
will relieve this problem to some extent. In other words, a model
with DBN is easier to optimize.  We validate this on the
\emph{S-plain} architecture with feature maps of dimension $d=48$ and
number of layers 20, 32 and 44.  The models are trained with a
mini-batch size of 128, momentum of 0.9 and weight decay of 0.0005.
We set the initial learning rate to be 0.1, dividing it by 5 at 80 and
120 epochs, and end training at 160 epochs. The results in Figure
\ref{fig:exp_deeper} (a) show that, with increased depth, the model
with BN was more difficult to optimize than with DBN.  We conjecture
that the approximate dynamical isometry of DBN alleviates this
problem.

\paragraph{Higher Learning Rate}

A network with Batch Normalization can benefit from high learning
rates, and thus faster training, because it reduces internal covariate
shift \cite{2015_ICML_Ioffe}.  Here, we show that DBN can help even
more.  We train the networks with BN and DBN with a $4\times$ higher
learning rate --- 0.4.  We use the \emph{S-plain} architecture with
feature maps of dimensions $d=42$ and divide the learning rate by 5 at
60, 100, 140, and 180 epochs.  The results in Figure
\ref{fig:exp_deeper} (b) show that DBN has significantly better
training accuracy than BN. We argue that DBN benefits from higher
learning rates because of its property of improved conditioning.

\subsection{Applying DBN to Residual Network in Practice}

Due to our current un-optimized implementation of DBN, it would incur
a high computational cost to replace all BN modules of a network with
DBN\footnote{See the appendix for more details on computational cost.}. We instead only decorrelate the activations among a subset of
layers. We find that this in practice is already effective for
residual networks \cite{2015_CVPR_He}, because the information in
previous layers can pass directly to the later layers through the
identity connections. We also show that we can improve upon residual
networks \cite{2015_CVPR_He,2016_CoRR_He,2016_CoRR_Zagoruyko} by using
only one DBN module before the first residual block, which introduces
negligible computation cost. In principle, an optimized implementation
of DBN will be much faster, and could be injected in multiple places
in the network with little overhead. However, optimizing the
implementation of DBN is beyond the scope of this work.

\paragraph{Residual Network on CIFAR-10}
We apply our method on residual networks \cite{2015_CVPR_He} by using
only one DBN module before the first residual block (denoted as
DBN-L1). We also consider DBN with adjustable scale (denoted as
DBN-scale-L1) as discussed in Section \ref{sec:trainingAndInfer}.  We
adopt the Torch implementation of residual
networks\footnote{https://github.com/facebook/fb.resnet.torch} and
follow the same experimental protocol as described in
\cite{2015_CVPR_He}.  We train the residual networks with depth 20,
32, 44 and 56 on CIFAR-10.  Table \ref{table:resnet1} shows the test
errors of these networks. Our methods obtain lower test errors
compared to BN over all 4 networks, and the improvement is more
dramatic for deeper networks.  Also, we see that DBN-scale-L1
marginally outperforms DBN-L1 in all cases. Therefore, we focus on
comparing DBN-scale-L1 to BN in later experiments.

\paragraph{Wide Residual Network on CIFAR}

We apply DBN to Wide Residual Network (WRN) \cite{2016_CoRR_Zagoruyko}
to improve the performance on CIFAR-10 and CIFAR-100.
Following the convention set in \cite{2016_CoRR_Zagoruyko}, we use the
abbreviation WRN-d-k to indicate a WRN with depth \emph{d} and width
\emph{k}.
We again adopt the publicly available Torch
implementation\footnote{https://github.com/szagoruyko/wide-residual-networks}
and follow the same setup as in \cite{2016_CoRR_Zagoruyko}.
The results in Table \ref{table:wr} show that DBN improves 
the original WRN on both datasets and both networks. In particular, we
reduce the test error by $3.74 \%$ and $18.27 \%$ on CIFAR-10 and
CIFAR-100, respectively.

\paragraph{Residual Network on ILSVRC-2012}
We further validate the scalability of our method on ILSVRC-2012
with 1000 classes ~\cite{2009_ImageNet}.
We use the given official 1.28M training images as a training set, and
evaluate the top-1 and top-5 classification errors on the validation
set with 50k images.
We use the 18, 34, 50 and 101-layer residual network (Res-18, Res-34, Res-50 and Res-101)
and perform single model and single-crop testing.
We follow the same experimental setup as described in
~\cite{2015_CVPR_He}, except that we use 4 GPUs instead of 8 for training Res-50 and 2 GPUs for training Res-18 and Res-34 (whose single crop results have not been previously reported): we apply SGD with a mini-batch size of 256 over
4 GPUs for Res-50 and 8 GPUs for Res-101, momentum of 0.9 and weight decay of 0.0001; we set the initial
learning rate of 0.1, dividing it by 10 at 30 and 60 epochs, and end
the training at 90 epochs. The results are shown in Table
\ref{table:ImageNet}. We can see that the DBN-scale-L1 achieves lower
test errors compared to the original residual networks.

\section{Conclusions}

In this paper, we propose Decorrelated Batch Normalization (DBN),
which extends Batch Normalization to include whitening over mini-batch
data.  We find that PCA whitening can sometimes be detrimental to
training because it causes \emph{stochastic axis swapping}, and
demonstrate that it is critical to use ZCA whitening, which avoids
this issue.
DBN retains the advantages of Batch Normalization while using
decorrelated representations to further improve models' optimization
efficiency and generalization abilities. This is because DBN can
maintain approximate dynamical isometry and improve the conditioning
of the Fisher Information Matrix. These properties are experimentally
validated, suggesting DBN has great potential to be used in designing
DNN architectures.

\vspace{0.15in}
\noindent\textbf{Acknowledgement}
This work is partially supported  by China Scholarship Council, NSFC-61370125 and SKLSDE-2017ZX-03. We also thank Jonathan Stroud and Lanlan Liu for their help with proofreading and editing. 

\setcounter{equation}{0}
\setcounter{figure}{0}
\setcounter{table}{0}
\setcounter{section}{0}
\renewcommand{\theequation}{\arabic{equation}}
\renewcommand{\thefigure}{\arabic{figure}}
\renewcommand{\thetable}{\arabic{table}}
\renewcommand{\thesection}{\Alph{section}}
\numberwithin{equation}{section}
\numberwithin{table}{section}
\numberwithin{figure}{section}
\section*{\Large{Appendix}}

\section{Derivation for Back-propagation}
\label{Sec_simpleForm_supp}
For illustration, we first provide the forward pass, then derive the backward pass. Regarding notation, we follow the matrix notation that all the vectors are  column vectors, except that the gradient vectors are  row vectors.

\subsection{Forward Pass}
Given mini-batch layer inputs  $\{\mathbf{x}_{i}, i=1,2...,m \}$ where \textit{m} is the number of examples, the ZCA-whitened output $\hat{\mathbf{x}}_i$ for the input $\mathbf{x}_i$ can be calculated as follows:
\begin{eqnarray}
&\mathbf{\mu}&=\frac{1}{m} \sum_{j=1}^{m} \mathbf{x}_j\\
&\Sigma &=\frac{1}{m} \sum_{j=1}^{m} (\mathbf{x}_j-\mathbf{\mu})(\mathbf{x}_j-\mathbf{\mu})^T \\
\label{equ_svd_supp}
&\Sigma &= \mathbf{D} \Lambda \mathbf{D}^T\\
\label{equ_project_supp}
&\mathbf{U}&=  \Lambda^{-1/2} \mathbf{D}^T \\
\label{equ_hatX_supp}
&\tilde{\mathbf{x}}_i &=  \mathbf{U} (\mathbf{x}_i - \mathbf{\mu})\\
\label{equ_activation_supp}
&\hat{\mathbf{x}}_i& =\mathbf{D}  \tilde{\mathbf{x}}_i
\end{eqnarray}

where $\mathbf{\mu}$ and $\Sigma$ are the mean vector and the covariance matrix within the mini-batch data. Eqn.~\ref{equ_svd_supp} is the eigen decomposition where $\mathbf{D}^T \mathbf{D}=I$ and $\Lambda$ is a diagonal matrix where the diagonal elements are the eigenvalues.
Note that $\tilde{\mathbf{x}}_i$ are auxiliary variables for clarity. Actually $\tilde{\mathbf{x}}_i$  are the output of PCA whitening. However, PCA whitening hardly works for deep networks as discussed in the paper.

\subsection{Back-propagation}

 Based on the chain rule and the result from \cite{2015_ICCV_Ionescu}, we can get the backward pass derivatives as follows:
\begin{align}
\label{eq_dLdoutput_supp}
  \frac{\partial L}{\partial \tilde{\mathbf{x}}_i}	&=\frac{\partial L}{\partial \hat{\mathbf{x}}_i}     \mathbf{D}\\
\label{eq_dLdU_supp}
  \frac{\partial L}{\partial \mathbf{U}}&= \sum_{i=1}^{m}	 \frac{\partial L}{\partial \tilde{\mathbf{x}}_i}^T (\mathbf{x}_i - \mathbf{\mu})^T \\
\label{eq_dLdLambda_supp}
  \frac{\partial L}{\partial \Lambda}&=(\frac{\partial L}{\partial \mathbf{U}}) \mathbf{D} (-\frac{1}{2} \Lambda^{-3/2} )\\
\label{eq_dLdD_supp}
  \frac{\partial L}{\partial \mathbf{D}}&=   \frac{\partial L}{\partial \mathbf{U}}^T \Lambda^{-1/2}
+\sum_{i=1}^{m} \frac{\partial L}{\partial \hat{\mathbf{x}}_i}^T \tilde{\mathbf{x}}_i^T  \\
\label{eq_dLdSigma_supp}
  \frac{\partial L}{\partial \Sigma}&=\mathbf{D}\{ (\mathbf{K}^T \odot (\mathbf{D}^T \frac{\partial L}{\partial \mathbf{D}} )) + (\frac{\partial L}{\partial \Lambda})_{diag}\} \mathbf{D}^T\\
\label{eq_dLdc_supp}
  \frac{\partial L}{\partial \mathbf{\mu}}&=\sum_{i=1}^{m}	\frac{\partial L}{\partial \tilde{\mathbf{x}}_i} (-\mathbf{U})+\sum_{i=1}^{m} \frac{-2(\mathbf{x}_i - \mathbf{\mu})^T}{m} (\frac{\partial L}{\partial \Sigma})_{sym}\\
\label{eq_dLdx_supp}
  \frac{\partial L}{\partial \mathbf{x}_i}&=\frac{\partial L}{\partial \tilde{\mathbf{x}}_i} \mathbf{U} + \frac{2(\mathbf{x}_i - \mathbf{\mu})^T}{m} (\frac{\partial L}{\partial \Sigma})_{sym} +\frac{1}{m} \frac{\partial L}{\partial \mathbf{\mu}}
\end{align}
where  $L$ is the loss function, $\mathbf{K} \in \mathbb{R}^{d\times d}$ is 0-diagonal with
$\mathbf{K}_{ij}= \frac{1}{\sigma_i  -\sigma_j} [i\neq j]$,
the $\odot$ operator is element-wise matrix multiplication, $(\frac{\partial
  L}{\partial \Lambda})_{diag}$ sets the off-diagonal elements of $\frac{\partial
  L}{\partial \Lambda}$ to zero, and $(\frac{\partial L}{\partial \Sigma})_{sym}$ means symmetrizing $\frac{\partial L}{\partial \Sigma}$ by $(\frac{\partial L}{\partial \Sigma})_{sym}=\frac{1}{2} (\frac{\partial L}{\partial \Sigma}^T+\frac{\partial L}{\partial \Sigma})$.
Note that Eqn.~\ref{eq_dLdSigma_supp} is from the results in \cite{2015_ICCV_Ionescu}. Besides, a similar formulation to  back-propagate the gradient through the whitening transformation has been derived in the context of learning an orthogonal weight matrix in \cite{2017_Huang_OWN}.

\subsection{Derivation for Simplified Formulation}
For more efficient computation, we provide the simplified formulation as follows:
 \begin{align}
 \label{equ_backPro_supp}
&\frac{\partial L}{\partial \mathbf{x}_i}&=
    (\frac{\partial L}{\partial \tilde{\mathbf{x}}_i} - \textbf{f} + \tilde{\mathbf{x}}_i^T \mathbf{S}  -    \tilde{\mathbf{x}}_i^T \mathbf{M}  ) \Lambda^{-1/2} \mathbf{D}^T,
\end{align}
where
\begin{align*}
\textbf{f}&=\frac{1}{m} \sum_{i=1}^{m} \frac{\partial L}{\partial \tilde{\mathbf{x}}_i} , \\
 \mathbf{S}&=2(\mathbf{K}^T \odot   (\Lambda \mathbf{F}_c^T +\Lambda^{\frac{1}{2}} \mathbf{F}_c \Lambda^{\frac{1}{2}}))_{sym}, \\
\mathbf{M}&= (\mathbf{F}_c)_{diag} \quad\mathrm{and}\quad \mathbf{F}_c=\frac{1}{m} (\sum_{i=1}^{m} \frac{\partial L}{\partial \tilde{\mathbf{x}}_i}^T \tilde{\mathbf{x}}_i^T).
\end{align*}

The details of how to derive Eqn. \ref{equ_backPro_supp} are as follows.

Based on Eqn. \ref{equ_project_supp}, \ref{equ_hatX_supp}, \ref{eq_dLdU_supp} and \ref{eq_dLdLambda_supp}, we can get
\begin{align}
	\frac{\partial L}{\partial \Lambda}=&(\sum_{i=1}^{m} \frac{\partial L}{\partial \tilde{\mathbf{x}}_i}^T	 (\mathbf{x}_i - \mathbf{\mu})^T ) \mathbf{U} (-\frac{1}{2} \Lambda^{-1} )	\nonumber \\
	=&-\frac{1}{2}(\sum_{i=1}^{m} \frac{\partial L}{\partial \tilde{\mathbf{x}}_i}^T \tilde{\mathbf{x}}_i^T )  \Lambda^{-1}.
\end{align}
Denoting $\mathbf{F}_c=\frac{1}{m} (\sum_{i=1}^{m} \frac{\partial L}{\partial \tilde{\mathbf{x}}_i}^T \tilde{\mathbf{x}}_i^T) $, we have
\begin{align}
	\frac{\partial L}{\partial \Lambda}&= -\frac{1}{2} m \mathbf{F}_c  \Lambda^{-1}.
\end{align}
Based on Eqn. \ref{equ_project_supp} and \ref{eq_dLdD_supp}, we have:
\begin{align}
	\mathbf{D}^T \frac{\partial L}{\partial \mathbf{D}}=& \Lambda^{\frac{1}{2}}\mathbf{U}^T (\frac{\partial L}{\partial U}^T \Lambda^{-\frac{1}{2}}+\sum_{i=1}^{m} \mathbf{D} \frac{\partial L}{\partial \tilde{\mathbf{x}}_i}^T \tilde{\mathbf{x}}_i^T)  \nonumber \\
	=&  m \Lambda^{\frac{1}{2}} \mathbf{F}_c^T \Lambda^{-\frac{1}{2}}+m \mathbf{F}_c.
\end{align}
Based on Eqn. \ref{eq_dLdSigma_supp}, we have :
\begin{align}
\dfrac{\partial L}{\partial \Sigma}&=\frac{1}{2}\mathbf{D} \{ (\mathbf{K}^T \odot (\mathbf{D}^T \dfrac{\partial L}{\partial \mathbf{D}} ))
  + (\mathbf{K} \odot (\mathbf{D}^T \dfrac{\partial L}{\partial \mathbf{D}})^T) \} \mathbf{D}^T \nonumber\\
  & \quad+ \mathbf{D} (\dfrac{\partial L}{\partial \Lambda})_{diag} \mathbf{D}^T \nonumber \\
&= \frac{1}{2}\mathbf{D} \{ \mathbf{K}^T \odot  m (\Lambda^{\frac{1}{2}} \mathbf{F}_c^T \Lambda^{-\frac{1}{2}} +\mathbf{F}_c ) \nonumber\\
   & \quad+ \mathbf{K} \odot  m (\Lambda^{-\frac{1}{2}} \mathbf{F}_c \Lambda^{\frac{1}{2}}+ \mathbf{F}_c^T ) \} \mathbf{D}^T
   + \mathbf{D} (\dfrac{\partial L}{\partial \Lambda})_{diag} \mathbf{D}^T \nonumber \\
 &= \frac{m}{2}\mathbf{D}  (\mathbf{K}^T \odot ( \Lambda^{\frac{1}{2}} \mathbf{F}_c^T \Lambda^{-\frac{1}{2}}+\mathbf{F}_c) )\mathbf{D}^T \nonumber\\
   &\quad +\frac{m}{2}\mathbf{D}(\mathbf{K} \odot  (\Lambda^{-\frac{1}{2}} \mathbf{F}_c \Lambda^{\frac{1}{2}}+\mathbf{F}_c^T) )  \mathbf{D}^T
   + \mathbf{D} (\dfrac{\partial L}{\partial \Lambda})_{diag} \mathbf{D}^T \nonumber \\
&=  \frac{m}{2}\mathbf{U}^T  (\mathbf{K}^T \odot  (\Lambda \mathbf{F}_c^T +\Lambda^{\frac{1}{2}} \mathbf{F}_c \Lambda^{\frac{1}{2}}) )\mathbf{U} \nonumber \\
   &\quad+  \frac{m}{2}\mathbf{U}^T(\mathbf{K} \odot  ( \mathbf{F}_c \Lambda+\Lambda^{\frac{1}{2}} \mathbf{F}_c^T \Lambda^{\frac{1}{2}}) )  \mathbf{U}
   + \mathbf{D} (\dfrac{\partial L}{\partial \Lambda})_{diag} \mathbf{D}^T \nonumber \\
&=  \frac{m}{2}\mathbf{U}^T \{ \mathbf{K}^T \odot   (\Lambda \mathbf{F}_c^T +\Lambda^{\frac{1}{2}} \mathbf{F}_c \Lambda^{\frac{1}{2}}) \nonumber\\
   &\quad+  \mathbf{K} \odot   ( \mathbf{F}_c \Lambda+\Lambda^{\frac{1}{2}} \mathbf{F}_c^T \Lambda^{\frac{1}{2}}) -(\Lambda^{\frac{1}{2}} \mathbf{F}_c \Lambda^{-\frac{1}{2}})_{diag} \} \mathbf{U} \nonumber \\
&=  \frac{m}{2}\mathbf{U}^T \{ 2(\mathbf{K}^T \odot   (\Lambda \mathbf{F}_c^T +\Lambda^{\frac{1}{2}} \mathbf{F}_c \Lambda^{\frac{1}{2}}))_{sym} \nonumber\\
   &\quad-(\Lambda^{\frac{1}{2}} \mathbf{F}_c \Lambda^{-\frac{1}{2}})_{diag} \} \mathbf{U}. \nonumber \\
&=  \frac{m}{2}\mathbf{U}^T \{ \mathbf{S}-\mathbf{M} \} \mathbf{U},
\end{align}
where  $\mathbf{S}=2(\mathbf{K}^T \odot   (\Lambda \mathbf{F}_c^T +\Lambda^{\frac{1}{2}} \mathbf{F}_c \Lambda^{\frac{1}{2}}))_{sym}$ and $\mathbf{M}= (\Lambda^{\frac{1}{2}} \mathbf{F}_c \Lambda^{-\frac{1}{2}})_{diag}=(\mathbf{F}_c)_{diag}$.
 Based on Eqn. \ref{eq_dLdc_supp}, we have
 \begin{align}
	\frac{\partial L}{\partial \mathbf{\mu}}&=(\sum_{i=1}^{m}	\frac{\partial L}{\partial \tilde{\mathbf{x}}_i}) (-\mathbf{U})+	 (\sum_{i=1}^{m} \frac{-2(\mathbf{x}_i - \mathbf{\mu})^T}{m}) (\frac{\partial L}{\partial \Sigma})_{sym} \nonumber \\
	&=-m \mathbf{f} \mathbf{U}
\end{align}
where $\mathbf{f}=\frac{1}{m} \sum_{i=1}^{m} \frac{\partial L}{\partial \tilde{\mathbf{x}}_i} $. We thus have:
\begin{align}
 \frac{\partial L}{\partial \mathbf{x}_i}&= \frac{\partial L}{\partial \tilde{\mathbf{x}}_i} \mathbf{U} + \frac{2(\mathbf{x}_i - \mathbf{\mu})^T}{m} (\frac{\partial L}{\partial \Sigma})_{sym}  +\frac{1}{m} \frac{\partial L}{\partial \mathbf{\mu}}  \nonumber \\
   &=\frac{\partial L}{\partial \tilde{\mathbf{x}}_i} \mathbf{U} +\tilde{\mathbf{x}}_i^T \mathbf{S} \mathbf{U}  - \tilde{\mathbf{x}}_i^T \mathbf{M} \mathbf{U} -\mathbf{f} \mathbf{U} \nonumber \\
   &=(\frac{\partial L}{\partial \tilde{\mathbf{x}}_i} - \mathbf{f} + \tilde{\mathbf{x}}_i^T \mathbf{S}  - \tilde{\mathbf{x}}_i^T \mathbf{M}  ) \mathbf{U}  \nonumber \\
    &=(\frac{\partial L}{\partial \tilde{\mathbf{x}}_i} - \mathbf{f} + \tilde{\mathbf{x}}_i^T \mathbf{S}  - \tilde{\mathbf{x}}_i^T \mathbf{M}  ) \Lambda^{-1/2} \mathbf{D}^T
\end{align}

\section{Computational Cost of DBN}
In this part, we analyze the computational cost of DBN module for Convolutional Neural Networks (CNNs).
Theoretically, a convolutional layer with a $d \times w\times
h$ input, batch size of $m$, and $d$ filters of size $F_h\times F_w$
costs $O(d^2 m h w F_h F_w)$. Adding DBN with a group size $K$ incurs
an overhead of $O(d K m h w + d K^2)$. The relative overhead is $\frac{K}{d F_h
  F_w}+\frac{K^2}{d m h w F_h F_w}$, which is negligible when $K$ is
small (\eg 16).

Empirically, our unoptimized implementation of  DBN
costs $71$ms (forward pass + backward pass, averaged over $10$ runs) for \emph{full} whitening, with a $64\times 32 \times 32$ input, a batch
size of $64$. In comparison, the highly optimized $3\times 3$ cudnn
convolution \cite{2014_cudnn} in Torch \cite{2011_torch} with the same input costs $32$ms.

\begin{table}[htb]
\caption{Time costs (s/per epoch) on VGG-A and CIFAR datasets with different groups.}
\label{DBN_Table1}
\begin{center}
\begin{small}
\begin{tabular}{lc|r}
\hline
Methods & Training  &  Inference   \\
\hline
BN &   163.68 &  11.47 \\
DBN-G8 & 707.47 &  15.50 \\
DBN-G16 & 466.92 &  14.41 \\
DBN-G64& 297.25 &  13.70 \\
DBN-G256 & 440.88 &  13.64 \\
DBN-G512 & 1004 &  13.68 \\

\hline
\end{tabular}
\vspace{-0.2in}
\end{small}
\end{center}
\end{table}

\begin{table}[htb]
\caption{Time costs (s/per epoch) for residual networks and wide residual network on CIFAR.}
  \label{DBN_Table2}
  \vspace{0.1in}
  \centering
  \begin{small}
  \begin{tabular}{c|cc|cc}

     \hline
    & \multicolumn{2}{c|}{Training} &    \multicolumn{2}{c}{Inference}   \\
    Method   & BN    & DBN-scale  & BN   & DBN-scale \\
  \hline
     Res-56   &69.53      & 86.80       &   4.57       & 5.12 \\
     Res-44   & 55.03       & 72.06      &   3.65         & 4.36 \\
     Res-32   & 40.36      & 57.47     &  2.80        & 3.33 \\
     Res-20   & 25.97      & 42.87     &   1.94         & 2.44 \\
     WideRes-40    &643.94      & 659.55    &    25.69       & 26.08  \\
     WideRes-28    & 440 & 457   &36.56   & 38.10 \\

     \hline
  \end{tabular}

\end{small}
\end{table}

\begin{table}[htb]
\caption{Time costs (s/per iteration) for residual networks on ImageNet, averaged 10 iterations, with multiple GPUs.}
  \label{DBN_Table3}
  \vspace{0.1in}
  \centering
  \begin{small}
  \begin{tabular}{c|cc|cc}

      \hline
    & \multicolumn{2}{c|}{Training} &    \multicolumn{2}{c}{Inference}   \\
    Method   & BN    & DBN-scale  & BN   & DBN-scale \\
  \hline
     Res-101  & 1.52      & 4.21     &   0.42      & 0.57 \\
     Res-50   & 0.51      & 1.12     &   0.19      & 0.25 \\
     Res-34   & 1.21      & 2.04     &   0.55      & 0.71 \\
     Res-18   & 0.81      & 1.45     &   0.40      & 0.51 \\

          \hline
  \end{tabular}

\end{small}
\end{table}

Table \ref{DBN_Table1}, \ref{DBN_Table2}, \ref{DBN_Table3} show the wall clock time for
our CIFAR-10 and ImageNet experiments described in the paper.
Note that DBN with small groups (\eg G8) can cost more time
than larger groups due to our unoptimized implementation: for example,
we whiten each group sequentially instead of in parallel, because Torch does not yet provide an
easy way to use linear algebra library of CUDA in parallel.
Our current implementation of DBN has a low GPU utilization
  (\eg $20\%$-$40\%$ on  average), versus $95\%+$
for BN. Thus there is a lot of room for a more efficient implementation.

\begin{table} [htb]
\caption{Time costs (s/per iteration) for residual networks on ImageNet, on a single GPU with a batch size of 32, averaged 10 iterations.}
\label{DBN_Table4}
\vspace{0.1in}
\centering
  \begin{small}
  \begin{tabular}{c|cc|cc}

      \hline
    & \multicolumn{2}{c|}{Training} &    \multicolumn{2}{c}{Inference}   \\
    Method   & BN    & DBN-scale  & BN   & DBN-scale \\
  \hline
     Res-101  & 1.24      & 1.35     &   0.35      & 0.37 \\
     Res-50   & 0.69      & 0.80     &   0.19      & 0.22 \\
     Res-34   & 0.34      & 0.45     &   0.12     & 0.14 \\
     Res-18   & 0.20      & 0.31     &   0.07      & 0.09 \\

          \hline
  \end{tabular}

\end{small}
\end{table}
We also observe that for the ResNet experiments on ImageNet, the overhead of multi-GPU parallelization
is relatively high in our current DBN implementation. Thus, we perform another set of ResNet experiments with the same settings except that we use a batch size of 32 and a single GPU.
As shown in Table \ref{DBN_Table4}, the difference in time cost between BN and DBN is much smaller on a single GPU. This suggests that there is room for optimizing our DBN implementation for multi-GPU training and inference.

{\small
\bibliographystyle{ieee}
\bibliography{whitening}

\begin{thebibliography}{10}\itemsep=-1pt

\bibitem{2016_ICML_Arpit}
D.~Arpit, Y.~Zhou, B.~U. Kota, and V.~Govindaraju.
\newblock Normalization propagation: {A} parametric technique for removing
  internal covariate shift in deep networks.
\newblock In {\em {ICML}}, volume~48 of {\em {JMLR} Workshop and Conference
  Proceedings}, pages 1168--1176. JMLR.org, 2016.

\bibitem{2016_CoRR_Ba}
L.~J. Ba, R.~Kiros, and G.~E. Hinton.
\newblock Layer normalization.
\newblock {\em CoRR}, abs/1607.06450, 2016.

\bibitem{1961_Barlow_possible}
H.~B. Barlow.
\newblock {\em {Possible principles underlying the transformations of sensory
  messages}}, pages 217--234.
\newblock MIT Press, Cambridge, MA, 1961.

\bibitem{1997_Vr_Bell}
A.~J. Bell and T.~J. Sejnowski.
\newblock {The "independent components" of natural scenes are edge filters.}
\newblock {\em Vision research}, 37(23):3327--3338, Dec. 1997.

\bibitem{2009_NIPS_Bengio}
Y.~Bengio and J.~S. Bergstra.
\newblock Slow, decorrelated features for pretraining complex cell-like
  networks.
\newblock In {\em Advances in Neural Information Processing Systems 22}, pages
  99--107. 2009.

\bibitem{CHHH07}
D.~Cai, X.~He, Y.~Hu, J.~Han, and T.~Huang.
\newblock Learning a spatially smooth subspace for face recognition.
\newblock In {\em Proc. IEEE Conf. Computer Vision and Pattern Recognition
  Machine Learning (CVPR'07)}, 2007.

\bibitem{2014_cudnn}
S.~Chetlur, C.~Woolley, P.~Vandermersch, J.~Cohen, J.~Tran, B.~Catanzaro, and
  E.~Shelhamer.
\newblock cudnn: Efficient primitives for deep learning.
\newblock {\em CoRR}, abs/1410.0759, 2014.

\bibitem{2017_Corr_Cho}
M.~Cho and J.~Lee.
\newblock Riemannian approach to batch normalization.
\newblock In I.~Guyon, U.~V. Luxburg, S.~Bengio, H.~Wallach, R.~Fergus,
  S.~Vishwanathan, and R.~Garnett, editors, {\em Advances in Neural Information
  Processing Systems 30}, pages 5225--5235. Curran Associates, Inc., 2017.

\bibitem{2015_CoRR_Clevert}
D.~Clevert, T.~Unterthiner, and S.~Hochreiter.
\newblock Fast and accurate deep network learning by exponential linear units
  (elus).
\newblock 2016.

\bibitem{2016_ICLR_Cogswell}
M.~Cogswell, F.~Ahmed, R.~B. Girshick, L.~Zitnick, and D.~Batra.
\newblock Reducing overfitting in deep networks by decorrelating
  representations.
\newblock In {\em ICLR}, 2016.

\bibitem{2011_torch}
R.~Collobert, K.~Kavukcuoglu, and C.~Farabet.
\newblock Torch7: A matlab-like environment for machine learning.
\newblock In {\em BigLearn, NIPS Workshop}, 2011.

\bibitem{2016_CoRR_Cooijmans}
T.~Cooijmans, N.~Ballas, C.~Laurent, and A.~C. Courville.
\newblock Recurrent batch normalization.
\newblock In {\em ICLR}, 2017.

\bibitem{2009_ImageNet}
J.~Deng, W.~Dong, R.~Socher, L.-J. Li, K.~Li, and L.~Fei-Fei.
\newblock {ImageNet: A Large-Scale Hierarchical Image Database}.
\newblock In {\em CVPR}, 2009.

\bibitem{2015_NIPS_Desjardins}
G.~Desjardins, K.~Simonyan, R.~Pascanu, and k.~kavukcuoglu.
\newblock Natural neural networks.
\newblock In {\em Advances in Neural Information Processing Systems 28}, pages
  2071--2079, 2015.

\bibitem{2015_NIPS_Guillaume}
G.~Desjardins, K.~Simonyan, R.~Pascanu, and K.~Kavukcuoglu.
\newblock Natural neural networks.
\newblock In {\em Proceedings of the 28th International Conference on Neural
  Information Processing Systems}, NIPS'15, pages 2071--2079, 2015.

\bibitem{2001_TPAMI_YaleB}
A.~Georghiades, P.~Belhumeur, and D.~Kriegman.
\newblock From few to many: Illumination cone models for face recognition under
  variable lighting and pose.
\newblock {\em IEEE Trans. Pattern Anal. Mach. Intelligence}, 23(6):643--660,
  2001.

\bibitem{2008_AD_Giles}
M.~B. Giles.
\newblock {\em Collected Matrix Derivative Results for Forward and Reverse Mode
  Algorithmic Differentiation}.
\newblock 2008.

\bibitem{2016_ICML_Grosse}
R.~B. Grosse and J.~Martens.
\newblock A kronecker-factored approximate fisher matrix for convolution
  layers.
\newblock In {\em {ICML}}, volume~48 of {\em {JMLR} Workshop and Conference
  Proceedings}, pages 573--582. JMLR.org, 2016.

\bibitem{2015_CVPR_He}
K.~He, X.~Zhang, S.~Ren, and J.~Sun.
\newblock Deep residual learning for image recognition.
\newblock In {\em arXiv prepring arXiv:1506.01497}, 2015.

\bibitem{2015_ICCV_He}
K.~He, X.~Zhang, S.~Ren, and J.~Sun.
\newblock Delving deep into rectifiers: Surpassing human-level performance on
  imagenet classification.
\newblock In {\em {ICCV}}. {IEEE} Computer Society, 2015.

\bibitem{2016_CoRR_He}
K.~He, X.~Zhang, S.~Ren, and J.~Sun.
\newblock Identity mappings in deep residual networks.
\newblock In B.~Leibe, J.~Matas, N.~Sebe, and M.~Welling, editors, {\em
  Computer Vision -- ECCV 2016}, pages 630--645, Cham, 2016. Springer
  International Publishing.

\bibitem{2016_CoRR_Huang_a}
G.~Huang, Z.~Liu, and K.~Q. Weinberger.
\newblock Densely connected convolutional networks.
\newblock {\em CoRR}, abs/1608.06993, 2016.

\bibitem{2017_Huang_PBWN}
L.~Huang, X.~Liu, B.~Lang, and B.~Li.
\newblock Projection based weight normalization for deep neural networks.
\newblock {\em CoRR}, abs/1710.02338, 2017.

\bibitem{2017_Huang_OWN}
L.~Huang, X.~Liu, B.~Lang, A.~W. Yu, Y.~Wang, and B.~Li.
\newblock Orthogonal weight normalization: Solution to optimization over
  multiple dependent stiefel manifolds in deep neural networks.
\newblock In {\em AAAI}, 2018.

\bibitem{Huang2017ICCV}
L.~Huang, X.~Liu, Y.~Liu, B.~Lang, and D.~Tao.
\newblock Centered weight normalization in accelerating training of deep neural
  networks.
\newblock In {\em ICCV}, 2017.

\bibitem{2017_NIPS_Ioffe}
S.~Ioffe.
\newblock Batch renormalization: Towards reducing minibatch dependence in
  batch-normalized models.
\newblock {\em CoRR}, abs/1702.03275, 2017.

\bibitem{2015_ICML_Ioffe}
S.~Ioffe and C.~Szegedy.
\newblock Batch normalization: Accelerating deep network training by reducing
  internal covariate shift.
\newblock In {\em Proceedings of the 32nd International Conference on Machine
  Learning, {ICML} 2015}, 2015.

\bibitem{2015_ICCV_Ionescu}
C.~Ionescu, O.~Vantzos, and C.~Sminchisescu.
\newblock Training deep networks with structured layers by matrix
  backpropagation.
\newblock In {\em Proceedings of International Conference on Computer Vision,
  {ICCV} 2015}, 2015.

\bibitem{2017_AS_Kessy}
A.~Kessy, A.~Lewin, and K.~Strimmer.
\newblock Optimal whitening and decorrelation.
\newblock {\em The American Statistician}, 0(ja):0--0, 2017.

\bibitem{2014_CoRR_Kingma}
D.~P. Kingma and J.~Ba.
\newblock Adam: {A} method for stochastic optimization.
\newblock {\em CoRR}, abs/1412.6980, 2014.

\bibitem{2009_TR_Alex}
A.~Krizhevsky.
\newblock Learning multiple layers of features from tiny images.
\newblock Technical report, 2009.

\bibitem{1992_WD_Krogh}
A.~Krogh and J.~A. Hertz.
\newblock A simple weight decay can improve generalization.
\newblock In {\em NIPS}. 1992.

\bibitem{2016_ICASSP_Laurent}
C.~Laurent, G.~Pereyra, P.~Brakel, Y.~Zhang, and Y.~Bengio.
\newblock Batch normalized recurrent neural networks.
\newblock In {\em 2016 {IEEE} International Conference on Acoustics, Speech and
  Signal Processing, {ICASSP} 2016}, pages 2657--2661, 2016.

\bibitem{1998_NN_Yann}
Y.~LeCun, L.~Bottou, G.~B. Orr, and K.-R. M\"{u}ller.
\newblock Effiicient backprop.
\newblock In {\em Neural Networks: Tricks of the Trade, This Book is an
  Outgrowth of a 1996 NIPS Workshop}, pages 9--50, 1998.

\bibitem{2016_axive_Liao}
Q.~Liao, K.~Kawaguchi, and T.~Poggio.
\newblock Streaming normalization: Towards simpler and more
  biologically-plausible normalizations for online and recurrent learning.
\newblock {\em arXiv preprint arXiv:1610.06160}, 2016.

\bibitem{2017_ICML_Luo}
P.~Luo.
\newblock Learning deep architectures via generalized whitened neural networks.
\newblock In {\em Proceedings of the 34th International Conference on Machine
  Learning}, pages 2238--2246, 2017.

\bibitem{2012_NN_Gregoire}
G.~Montavon and K.-R. M{\"u}ller.
\newblock {\em Deep {B}oltzmann Machines and the Centering Trick}, volume 7700
  of {\em LNCS}.
\newblock Springer, 2nd edn edition, 2012.

\bibitem{2010_ICML_Nair}
V.~Nair and G.~E. Hinton.
\newblock Rectified linear units improve restricted boltzmann machines.
\newblock In {\em Proceedings of the 27th International Conference on Machine
  Learning {ICML} 2010}, 2010.

\bibitem{2015_NIPS_Neyshabur}
B.~Neyshabur, R.~Salakhutdinov, and N.~Srebro.
\newblock Path-sgd: Path-normalized optimization in deep neural networks.
\newblock In {\em Annual Conference on Neural Information Processing Systems
  {NIPS} 2015}, pages 2422--2430, 2015.

\bibitem{2012_AISTATS_Raiko}
T.~Raiko, H.~Valpola, and Y.~LeCun.
\newblock Deep learning made easier by linear transformations in perceptrons.
\newblock In {\em International Conference on Artificial Intelligence and
  Statistics ({AISTATS})}, pages 924--932, 2012.

\bibitem{2017_ICLR_Ren}
M.~Ren, R.~Liao, R.~Urtasun, F.~H. Sinz, and R.~S. Zemel.
\newblock Normalizing the normalizers: Comparing and extending network
  normalization schemes.
\newblock 2017.

\bibitem{2017_ICLR_Pau}
P.~Rodr{\'{\i}}guez, J.~Gonz{\`{a}}lez, G.~Cucurull, J.~M. Gonfaus, and F.~X.
  Roca.
\newblock Regularizing cnns with locally constrained decorrelations.
\newblock 2017.

\bibitem{2016_CoRR_Salimans}
T.~Salimans and D.~P. Kingma.
\newblock Weight normalization: {A} simple reparameterization to accelerate
  training of deep neural networks.
\newblock {\em CoRR}, abs/1602.07868, 2016.

\bibitem{2013_CoRR_Saxe}
A.~M. Saxe, J.~L. McClelland, and S.~Ganguli.
\newblock Exact solutions to the nonlinear dynamics of learning in deep linear
  neural networks.
\newblock {\em CoRR}, abs/1312.6120, 2013.

\bibitem{1992_NC_Schmidhuber}
J.~Schmidhuber.
\newblock Learning factorial codes by predictability minimization.
\newblock {\em Neural Computation}, 4(6):863--879, 1992.

\bibitem{1998_Schraudolph}
N.~N. Schraudolph.
\newblock Accelerated gradient descent by factor-centering decomposition.
\newblock Technical report, 1998.

\bibitem{2002_FGR_Terence}
T.~Sim, S.~Baker, and M.~Bsat.
\newblock The cmu pose, illumination, and expression (pie) database.
\newblock In {\em Proceedings of the Fifth IEEE International Conference on
  Automatic Face and Gesture Recognition}, FGR '02, pages 53--, 2002.

\bibitem{2014_CoRR_Simonyan}
K.~Simonyan and A.~Zisserman.
\newblock Very deep convolutional networks for large-scale image recognition.
\newblock {\em CoRR}, abs/1409.1556, 2014.

\bibitem{2016_CoRR_Sun}
K.~Sun and F.~Nielsen.
\newblock Relative natural gradient for learning large complex models.
\newblock {\em CoRR}, abs/1606.06069, 2016.

\bibitem{2016_CoRR_Szegedy}
C.~Szegedy, S.~Ioffe, and V.~Vanhoucke.
\newblock Inception-v4, inception-resnet and the impact of residual connections
  on learning.
\newblock {\em CoRR}, abs/1602.07261, 2016.

\bibitem{2015_CoRR_Szegedy}
C.~Szegedy, V.~Vanhoucke, S.~Ioffe, J.~Shlens, and Z.~Wojna.
\newblock Rethinking the inception architecture for computer vision.
\newblock {\em CoRR}, abs/1512.00567, 2015.

\bibitem{2011_NIPS_Wiesler}
S.~Wiesler and H.~Ney.
\newblock A convergence analysis of log-linear training.
\newblock In {\em {NIPS}}, pages 657--665, 2011.

\bibitem{2014_ICASSP_Wiesler}
S.~Wiesler, A.~Richard, R.~Schl{\"{u}}ter, and H.~Ney.
\newblock Mean-normalized stochastic gradient for large-scale deep learning.
\newblock In {\em {ICASSP}}, pages 180--184. {IEEE}, 2014.

\bibitem{2017_Corr_Xiang}
S.~Xiang and H.~Li.
\newblock On the effects of batch and weight normalization in generative
  adversarial networks.
\newblock {\em CoRR}, abs/1704.03971, 2017.

\bibitem{2016_ICDM_Xiong}
W.~Xiong, B.~Du, L.~Zhang, R.~Hu, and D.~Tao.
\newblock Regularizing deep convolutional neural networks with a structured
  decorrelation constraint.
\newblock In {\em {IEEE} 16th International Conference on Data Mining, {ICDM}
  2016}, 2016.

\bibitem{2016_CoRR_Zagoruyko}
S.~Zagoruyko and N.~Komodakis.
\newblock Wide residual networks.
\newblock {\em CoRR}, abs/1605.07146, 2016.

\end{thebibliography}
}

\end{document}